\documentclass[10pt,twocolumn,letterpaper]{article}

\usepackage{cvpr}              

\usepackage{graphicx}
\usepackage{amsmath}
\usepackage{amssymb}
\usepackage{booktabs}

\usepackage[accsupp]{axessibility}  
\usepackage{algorithm}
\usepackage{algorithmic}
\usepackage{amsfonts,amssymb} 
\usepackage{amsmath}
\usepackage{color}
\usepackage{multirow}
\usepackage{pythonhighlight}
\usepackage{setspace}
\usepackage{amssymb}
\usepackage{multicol}
\usepackage{amssymb}
\usepackage{bbm}

\usepackage[pagebackref,breaklinks,colorlinks]{hyperref}

\usepackage[capitalize]{cleveref}
\crefname{section}{Sec.}{Secs.}
\Crefname{section}{Section}{Sections}
\Crefname{table}{Table}{Tables}
\crefname{table}{Tab.}{Tabs.}

\begin{document}

\title{Industrial Anomaly Detection with Domain Shift: A Real-world Dataset and Masked Multi-scale Reconstruction}

\author{Zilong~Zhang$^{1}$\qquad~Zhibin~Zhao$^{1}$\qquad~Xingwu~Zhang$^{1}$\thanks{Corresponding author}\qquad\\~Chuang~Sun$^{1}$\qquad~Xuefeng~Chen$^{1}$\\
		School of Mechanical Engineering, Xi’an Jiaotong University$^{1}$ \\
		{\tt\small zhangzilongc@gmail.com} \quad
		{\tt\small zhaozhibin@xjtu.edu.cn} \quad
            {\tt\small xwzhang@mail.xjtu.edu.cn} \quad\\
		{\tt\small ch.sun@xjtu.edu.cn} \quad
		{\tt\small chenxf@xjtu.edu.cn} \quad
	}

\maketitle

\begin{abstract}

Industrial anomaly detection (IAD) is crucial for automating industrial quality inspection. The diversity of the datasets is the foundation for developing comprehensive IAD algorithms. Existing IAD datasets focus on diversity of data categories, overlooking the diversity of domains within the same data category. In this paper, to bridge this gap, we propose the Aero-engine Blade Anomaly Detection (AeBAD) dataset, consisting of two sub-datasets: the single-blade dataset and the video anomaly detection dataset of blades. Compared to existing datasets, AeBAD has the following two characteristics: 1.) The target samples are not aligned and at different scales. 2.) There is a domain shift between the distribution of normal samples in the test set and the training set, where the domain shifts are mainly caused by the changes in illumination and view. Based on this dataset, we observe that current state-of-the-art (SOTA) IAD methods exhibit limitations when the domain of normal samples in the test set undergoes a shift. To address this issue, we propose a novel method called masked multi-scale reconstruction (MMR), which enhances the model's capacity to deduce causality among patches in normal samples by a masked reconstruction task. MMR achieves superior performance compared to SOTA methods on the AeBAD dataset. Furthermore, MMR achieves competitive performance with SOTA methods to detect the anomalies of different types on the MVTec AD dataset. Code and dataset are available at \href{https://github.com/zhangzilongc/MMR}{https://github.com/zhangzilongc/MMR}.

\end{abstract}

\section{Introduction}
\label{1_introduction}

Industrial anomaly detection (IAD) involves the identification and localization of anomalies in industrial processes, ranging from subtle changes like thin scratches to larger structural defects like missing components  \cite{bergmann2019mvtec}, with limited or no prior knowledge of abnormality. IAD has numerous applications, such as smart manufacturing processes that ensure the production of high-quality products and automated maintenance that relies on robots. Collecting sufficient abnormal samples for training is often difficult since industrial processes are generally optimized to minimize the production of defective products \cite{yang2020dfr}. Additionally, since industrial processes are affected by uncontrollable random factors, different types of abnormal samples may be produced \cite{zeiser2023evaluation, nag2022wafersegclassnet}. In other words, various and sufficient abnormal samples are generally difficult to obtain. Therefore, IAD is typically conducted in a one-class learning setting, where only normal \footnote{In this paper, normal is by definition an antonym of abnormal.} data is used.

\begin{figure*}[t]
\centering
\includegraphics[width=5.5in]{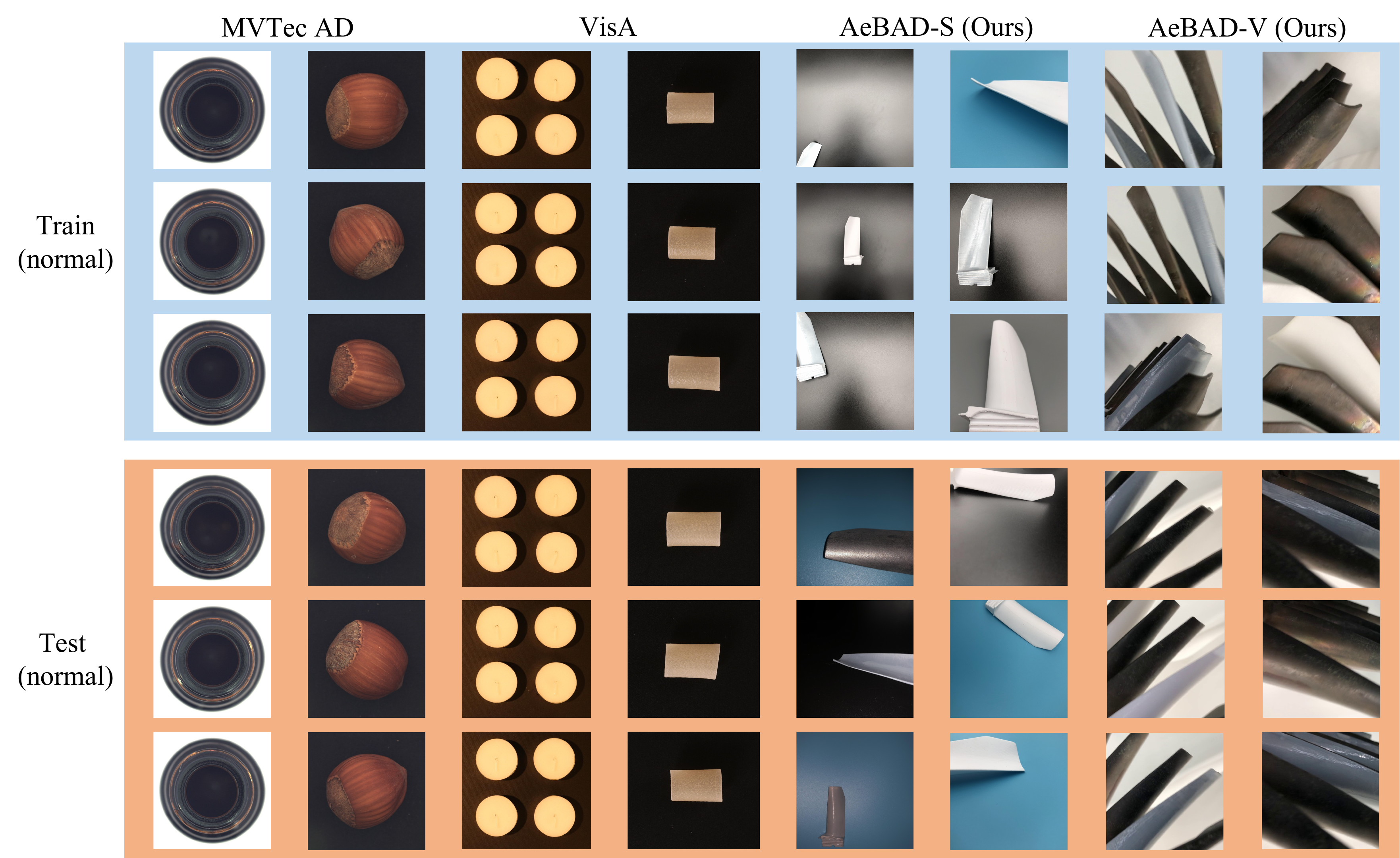}
\caption{Normal samples of the training set and the test set in the current datasets and our dataset. The characteristics of proposed AeBAD dataset are: 1.) The target samples are not aligned and at different scales. 2.) There is a domain shift between the distribution of normal samples in the test set and the training set, where the domain shifts are mainly caused by the changes in illumination and view. AeBAD dataset is developed for automatically detecting abnormalities of the blades of aero-engines to ensure the stable operation of the aero-engines.}
\label{fig_1}
\end{figure*}

Existing IAD methods \cite{roth2022towards, deng2022anomaly, li2021cutpaste, salehi2021multiresolution, you2022unified, schluter2022natural, zavrtanik2021draem} primarily rely on datasets such as MVTec AD \cite{bergmann2019mvtec}, VisA \cite{zou2022spot}, and MVTec LOCO AD \cite{bergmann2022beyond}, which share some common characteristics, such as aligned samples \footnote{Some images of other objects in the VISA dataset depict not registered objects.} and consistent scale within each class (as shown in Figure \ref{fig_1}). However, in real-world automated product/component inspection scenarios, these conditions may not always hold. Factors such as inaccurate control or environmental changes can cause significant variations in object scale and viewpoint \cite{9927464, 8963630}. This suggests that the mild conditions provided by existing datasets may not fully capture the complexities of real-world IAD applications. As a result, it is unclear whether current IAD methods can achieve the same level of success under such complex conditions, which raises important questions for the future development and evaluation of IAD techniques.

To address the aforementioned problem and bridge the gap between the real-world application of IAD and current datasets, we present the \textbf{A}ero-\textbf{e}ngine \textbf{B}lade \textbf{A}nomaly \textbf{D}etection (AeBAD) Dataset in this paper. The aim of AeBAD is to automatically detect abnormalities in the blades of aero-engines, ensuring their stable operation. Unlike previous datasets that focus on detecting diversity of defect categories, AeBAD is centered on the diversity of domains within the same data category. AeBAD consists of two sub-datasets: the single-blade dataset (AeBAD-S) and the video anomaly detection of blades (AeBAD-V). AeBAD-S comprises images of single blades of different scales, with a primary feature being that the samples are not aligned. Furthermore, there is a domain shift between the distribution of normal samples in the test set and the training set, where the domain shifts are mainly caused by the changes in illumination and view. Some examples are shown in Figure \ref{fig_1}. AeBAD-V, on the other hand, includes videos of blades assembled on the blisks of aero-engines, with the aim of detecting blade anomalies during blisk rotation. A distinctive feature of AeBAD-V is that the shooting view in the test set differs from that in the training set. The partial frames of videos are shown in Figure \ref{fig_1}. AeBAD-V is highly consistent with the real-world need for aero-engine blade detection, emphasizing the significance of IAD.

After constructing the AeBAD dataset, we evaluate the state-of-the-art (SOTA) IAD methods and make the following observations. Firstly, methods that rely on synthetic anomalies perform poorly when the samples are not aligned and at different scales, as most of the synthetic anomalies are distributed in non-target areas such as the background. This reduces the difficulty of discriminating normal samples from synthetic anomalies, resulting in poor generalization to unseen anomalies. Secondly, when the domain of normal samples in the test set shifts, the abnormal score of normal parts of the samples increases for all test methods, leading to a large number of false positives. This is because some current methods rely heavily on modeling the distribution of normal samples in the training set, which causes a mismatch between the features of the normal samples in the training set and those in the test set when the distribution changes. We believe that the root cause of this problem is the neglect of causality in normal samples.

Inspired by the above observations and analyses, we propose a method entitled \textbf{M}asked \textbf{M}ulti-scales \textbf{R}econstruction (MMR) for reconstructing multi-scale features of full images from masked inputs. Our method is free from synthetic anomalies and thus bypass the one of the above problems. Simultaneously, the pretext task of MMR enhances the model's perception of causality in normal samples, which improves the robustness to domain shifts. In contrast to existing inpainting methods, our method focuses on reconstructing features at different scales rather than tanglesome details, which provides the output with greater semantic clarity. Moreover, our masked strategy is different with the previous, where our masked parts are completely hidden. We will demonstrate that this masked strategy can significantly improve the performance of anomaly detection. Our experiments show that MMR outperforms the state-of-the-art (SOTA) methods in terms of reducing false positives and boosting anomaly detection performance on the AeBAD dataset. Additionally, MMR achieves comparable performance with SOTA methods on the MVTec AD dataset.

The main contributions of this paper are as follows:

\begin{enumerate}
	\item We recognize the gap between the actual application of IAD and the available datasets, and thus introduce a new dataset named AeBAD. Unlike previous datasets, AeBAD takes into account various practical factors such as different scales and views, which induces a domain shift for normal samples in the test set. This dataset closely aligns with the requirements for detecting anomalies in aero-engine blades and highlights the practical significance of IAD. To the best of our knowledge, this is the first industrial anomaly detection dataset that accounts for domain shifts, making it more representative of real-world scenarios.
	\item We observe that existing SOTA methods for IAD exhibit limitations when the domain of normal samples in the test set undergoes a shift. To address this issue, we propose a novel method called MMR, which enhances the model's capacity to deduce causality among patches in normal samples. Our method achieves superior performance compared to SOTA methods on the AeBAD dataset. Furthermore, MMR achieves competitive performance with SOTA methods on the MVTec AD dataset.
\end{enumerate}

The rest paper is organized as follows. Section \ref{2_related} reviews some related works. Section \ref{3_ae} describes our dataset and show the deficiency of current methods. The proposed MMR is given in Section \ref{4_method}. Section \ref{5_exp} presents the experimental results. Section \ref{6_abl} ablates the factors that are closely related to MMR. Section \ref{7_lim} presents the limitation of our work and the future work. A conclusion is presented in Section \ref{7_con}.

\section{Related Work}
\label{2_related}

We ﬁrst review existing datasets for IAD. Then, we give an overview of relevant approaches to IAD. More comprehensive surveys of IAD can be found in \cite{jiang2022survey, cui2022survey, tao2022deep}. In this paper, we do not review the methods and datasets related to semantic anomalies.

\subsection{Datasets}

The success of current deep learning methods heavily relies on the diversity of data distribution. This diversity encompasses a large number of data categories, such as ImageNet \cite{russakovsky2015imagenet}, as well as different domains of the same data category, such as NICO \cite{zhang2022nico++} and OOD-CV \cite{zhao2022ood}. Evaluating the performance of target algorithms on datasets with different domains is a crucial step in applying them in real-world scenarios. Despite the availability of several datasets containing various industrial products for the task of IAD, like MVTec AD \cite{bergmann2019mvtec}, DAGM \cite{wieler2007weakly}, KSDD \cite{tabernik2020segmentation} and SensumSODF \cite{ravcki2022detection}, to the best of our knowledge, all of them overlook the presence of domain shift between the training and test sets, which exists in crack segmentation \cite{liu2023learning}, railway track maintenance using robotic vision \cite{rahman2023railway}, surface defect detection \cite{shi2023few} and etc. Addressing this issue and considering domain shift is vital for developing more robust and effective IAD algorithms that can be reliably applied in practical scenarios.

Bergmann et al. \cite{bergmann2019mvtec} proposed the MVTec Anomaly Detection dataset (MVTec AD), which focuses on detecting subtle changes and larger structural defects. The main feature of MVTec AD is that all objects are roughly aligned, as shown in Figure \ref{fig_1}. Zou et al. \cite{zou2022spot} proposed the Visual Anomaly (VisA) Dataset, which considers more complex structures and presents multiple objects in a single image. The recent proposed MVTec Logical Constraints Anomaly Detection (MVTec LOCO AD) dataset \cite{bergmann2022beyond} aims to detect logical anomalies in addition to structural anomalies. Additionally, Bergmann et al. \cite{bergmann2021mvtec} introduced a 3D dataset for IAD.

However, all existing datasets ignore the presence of domain shift between the training and test sets. In this paper, inspired by the real-world application of aero-engine blade anomaly detection, we propose a novel dataset that includes domain shifts between the training and test sets. These domain shifts encompass different views and illuminations, and our proposed dataset can address some of the deficiencies of current datasets.

\subsection{Methods}

While there are different types of setups in IAD, such as normal data mixed with few noisy data (anomalies) \cite{xisoftpatch}, this paper focuses on practical applications and reviews only one-class learning in IAD, where only normal data is available in training. We categorize these methods into three categories based on the formulation of anomaly detection.

\begin{figure*}[!t]
\centering
\includegraphics[width=6in]{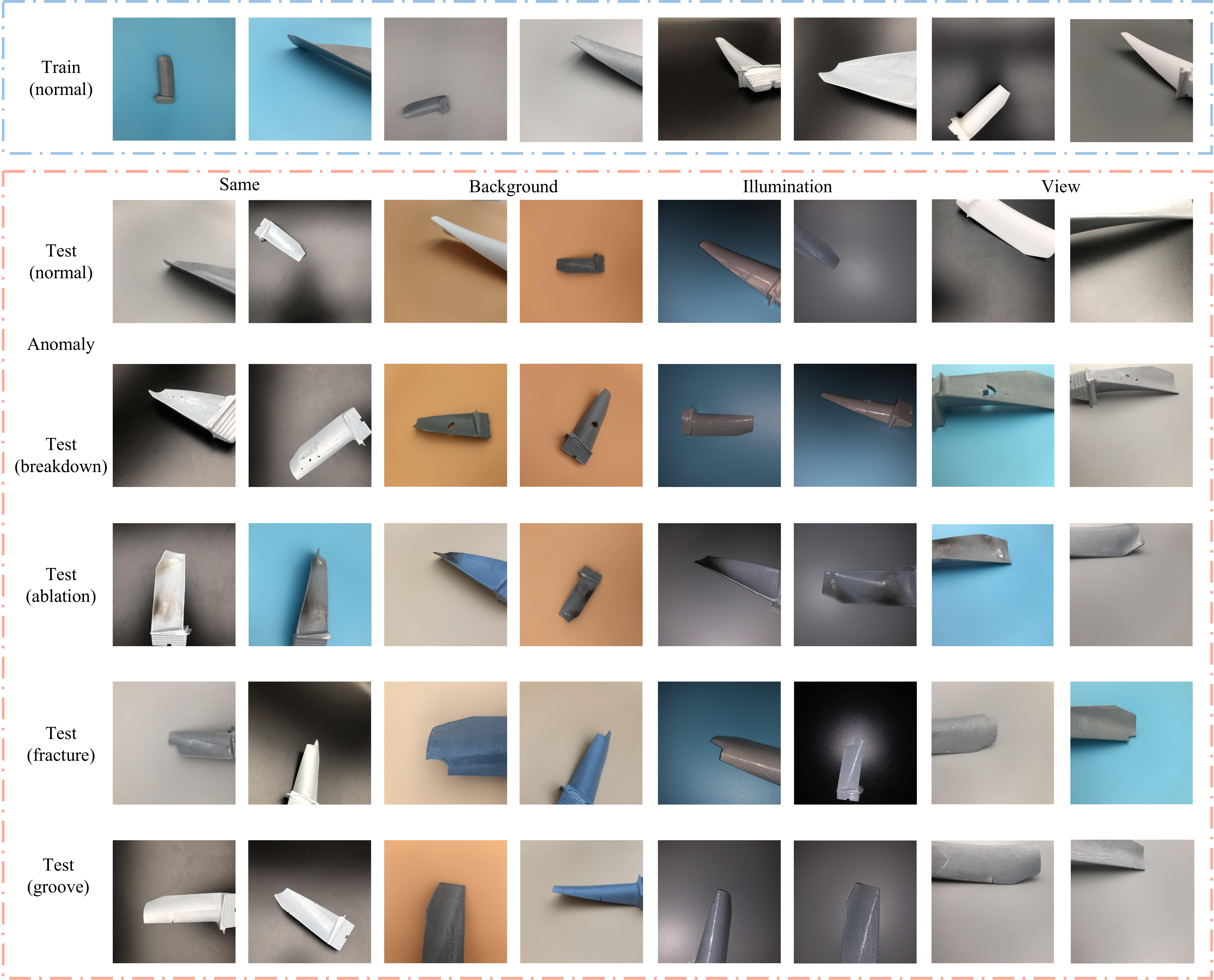}
\caption{The samples of the training set and the test set on AeBAD-S dataset. The training set consists solely of normal samples. The test set consists of four categories (every two in a row is a category), including same, background, illumination and view. In the "same" category, the distribution of the normal samples in the test set is the same as that in the training set. The "background" category involves changes to the image background, while the "illumination" category involves changes to the lighting conditions. The "view" category involves changes to the shooting positions, with the new positions being symmetrical to those in the training set. Additionally, the anomalies in the blade samples are divided into four categories: breakdown, ablation, fracture, and groove. The groove category refers to tiny defects on the margin of the blade, while the fracture category refers to missing areas of the blade. (\textbf{best view in color})}
\label{fig_3}
\end{figure*}

The first category formulates anomaly detection as a matching problem. These methods establish a pattern based on normal data and match the pattern of test data with the established pattern during testing. For example, PaDiM \cite{defard2020padim} establishes the distribution of every patch of normal data and calculates the likelihood of test data to detect anomalies. Similarly, FastFlow \cite{yu2021fastflow} and CFLOW-AD \cite{gudovskiy2022cflow} model the normalizing flows of normal data. PatchCore \cite{roth2022towards} builds a “database” of multi-scale features and matches the test data through k-nearest neighbor search. MKDAD \cite{salehi2021multiresolution} and ReverseKD \cite{deng2022anomaly} match the general pattern generated by the teacher network with the pattern of normal data generated by the student network to detect anomalies.

 The second category hopes the model can generalize the ability that distinguishes between normal and synthetic abnormalities to distinguish unseen abnormalities. The main difference among these methods is the construction of synthetic anomalies. For instance, CutPaste \cite{li2021cutpaste} crops the parts of a certain object into different objects to synthesize anomalies. NSA \cite{schluter2022natural} uses Poisson image editing to seamlessly clone an object from one image into another image, which can significantly reduce discontinuities. DRAEM \cite{zavrtanik2021draem} generates anomalies based on the mask generated by perlin noise and an additional dataset.

The third category includes methods based on reconstruction. These methods \cite{bergmann2018improving, dehaene2020iterative, matsubara2020deep} rely on the hypothesis that reconstruction models trained on normal samples only succeed in normal regions but fail in anomalous regions \cite{you2022unified}. The main problem with this idea is that the model could learn tricks that anomalies are also restored well. To tackle this problem, different strategies are carried out. RIAD \cite{zavrtanik2021reconstruction} and InTra \cite{pirnay2022inpainting} propose a masked image inpainting pretext task. Since some parts of the image are masked out, the model needs to perceive the causality in the image to recover the information of masked parts, which avoids the trivial solution. UniAD \cite{you2022unified} proposes a neighbor masked encoder and a layer-wise query decoder to prevent the "identical shortcut".

The methods most related to our work are RIAD \cite{zavrtanik2021reconstruction} and InTra \cite{pirnay2022inpainting}. The main differences are the following. First, we neglect to reconstruct intricate details in the full image and instead reconstruct features at different scales. This empowers the output of the model with more semantic information. Second, our masked strategy is different, where the masked parts are completely hidden, resulting in the complete removal of features in the masked parts, while in the previous methods, these features are shared with the features of the visible parts.

\begin{figure}[!t]
\centering
\includegraphics[width=2.5in]{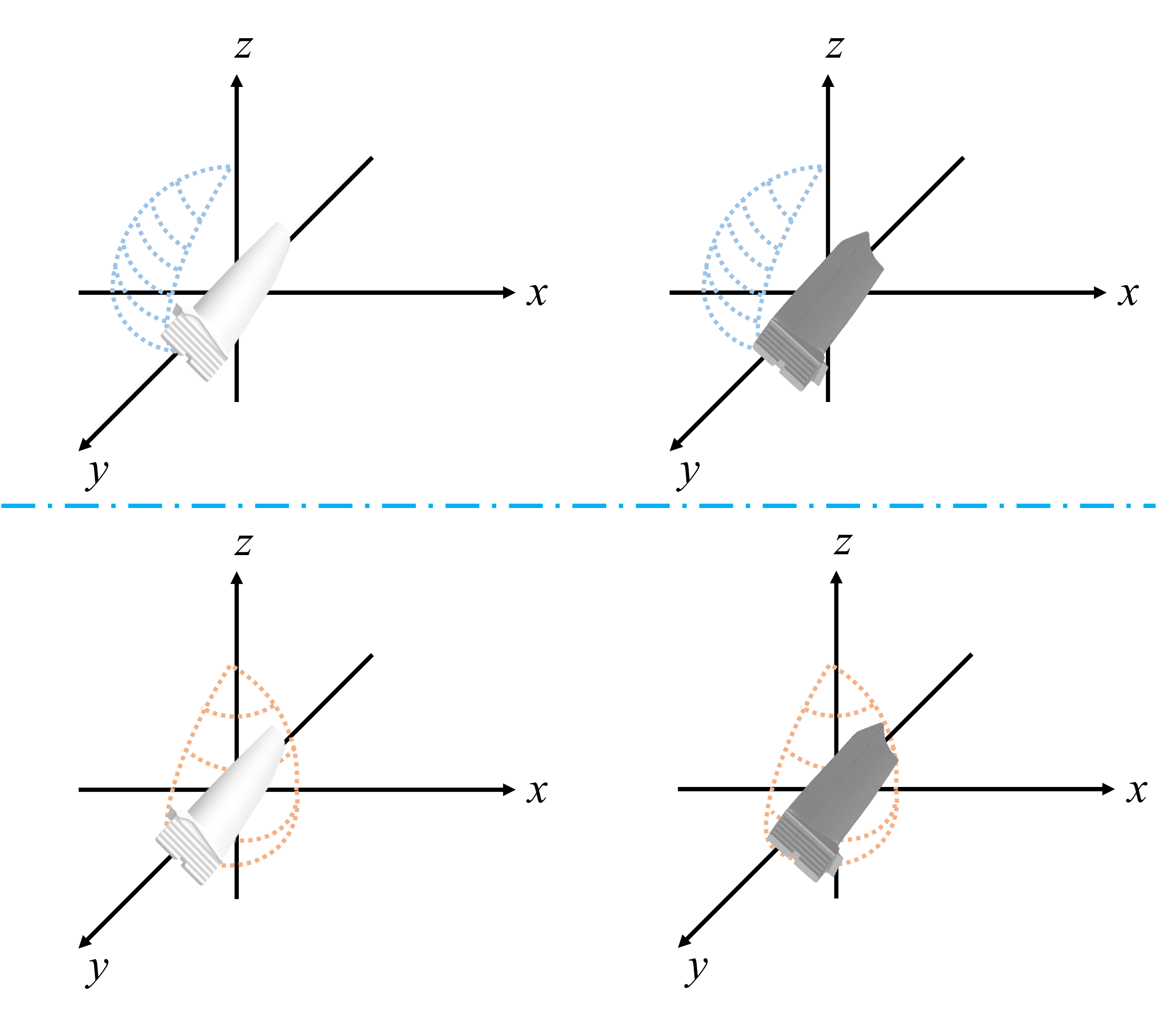}
\caption{\textbf{Top}: The shooting positions of the acquired images in the training set on AeBAD-S. All shooting positions are sampled from the positions of blue dotted lines. \textbf{Bottom}: The shooting positions of the acquired images in the “view” test set. All shooting positions (orange dotted lines) are sampled from the positions being symmetrical to the positions of blue dotted lines. (\textbf{best view in color})}
\label{fig_2}
\end{figure}

\section{Aero-engine Blade Anomaly Detection Dataset}
\label{3_ae}

\subsection{Background and Significance}

The aero-engine is the heart of an airplane, and its healthy blades are critical for ensuring stable operation. Defects in aero-engine blades can significantly impact the engine's efficiency or even cause it to fail \cite{9363206}. Due to their high cost and infrequent occurrence, obtaining a variety of blades with different defects (anomalies) can be difficult. As a result, blade defect detection is often approached as a one-class anomaly detection problem. In real-world blade detection scenarios, images of blades acquired by borehole instruments can vary in view and illumination, depending on the location of the borehole on the aero-engine. Since the aero-engine to be detected is unknown, the view is also unknown. This motivates us to explore one-class IAD under the domain shift of test set. For meeting the need of the real-world application, the blades in the experiment are either from the real engine or a 1:1 replica.

\begin{table}
\scriptsize
\centering
\caption{The statistical overview of AeBAD dataset. The videos in AeBAD-V have been divided into consecutive frames. "N" denotes the number of the normal samples, "A" denotes the number of the abnormal samples.}
	\begin{tabular}{|c|c|c|c|c|c|c|c|c|c|c|c|c|c|}
   \hline
 & Train & 
\multicolumn{12}{c|}{Test} \\ \hline
\multirow{3}{*}{AeBAD-S} &
\multirow{3}{*}{521} &
\multicolumn{3}{c|}{same} &
\multicolumn{3}{c|}{background} &
\multicolumn{3}{c|}{illumination} &
\multicolumn{3}{c|}{view} \\ \cline{3-14}
& & \multicolumn{2}{c|}{N} & \multicolumn{1}{c|}{A} & \multicolumn{2}{c|}{N} & \multicolumn{1}{c|}{A} & \multicolumn{2}{c|}{N} & \multicolumn{1}{c|}{A} & \multicolumn{2}{c|}{N} & \multicolumn{1}{c|}{A} \\ \cline{3-14}
& & \multicolumn{2}{c|}{230} & \multicolumn{1}{c|}{459} & \multicolumn{2}{c|}{93} & \multicolumn{1}{c|}{212} & \multicolumn{2}{c|}{75} & \multicolumn{1}{c|}{198} & \multicolumn{2}{c|}{92} & \multicolumn{1}{c|}{280} \\ \hline
\multirow{3}{*}{AeBAD-V} &
\multirow{3}{*}{707} &
\multicolumn{5}{c|}{video-1} &
\multicolumn{5}{c|}{video-2} &
\multicolumn{2}{c|}{video-3} \\ \cline{3-14}
& & \multicolumn{3}{c|}{N} & \multicolumn{2}{c|}{A} & \multicolumn{3}{c|}{N} & \multicolumn{2}{c|}{A} & \multicolumn{1}{c|}{N} & \multicolumn{1}{c|}{A} \\ \cline{3-14}
& & \multicolumn{3}{c|}{533} & \multicolumn{2}{c|}{368} & \multicolumn{3}{c|}{251} & \multicolumn{2}{c|}{650} & \multicolumn{1}{c|}{381} & \multicolumn{1}{c|}{520} \\ \hline
\end{tabular}
\label{Table_overview}
\end{table}

\begin{figure}[!t]
\centering
\includegraphics[width=3in]{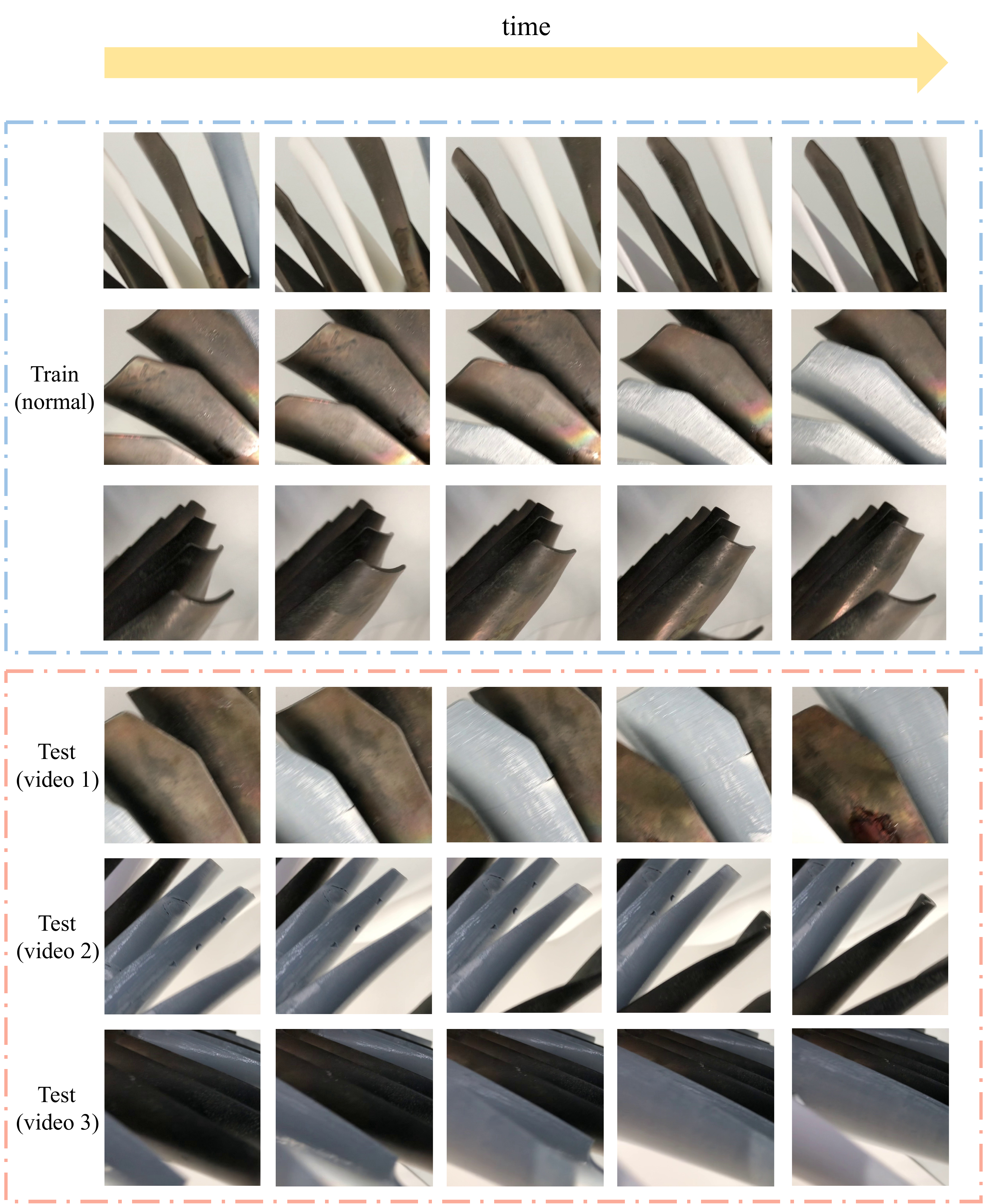}
\caption{The frames of videos of the training and the test set in AeBAD-V dataset. Each row, from left to right, is a sequence of frames over time. Different from AeBAD-S, there are multiple blades in one image. The shooting viewpoints of the test set are different from that of the training set.}
\label{fig_4}
\end{figure}

\begin{figure*}[!t]
\centering
\includegraphics[width=5in]{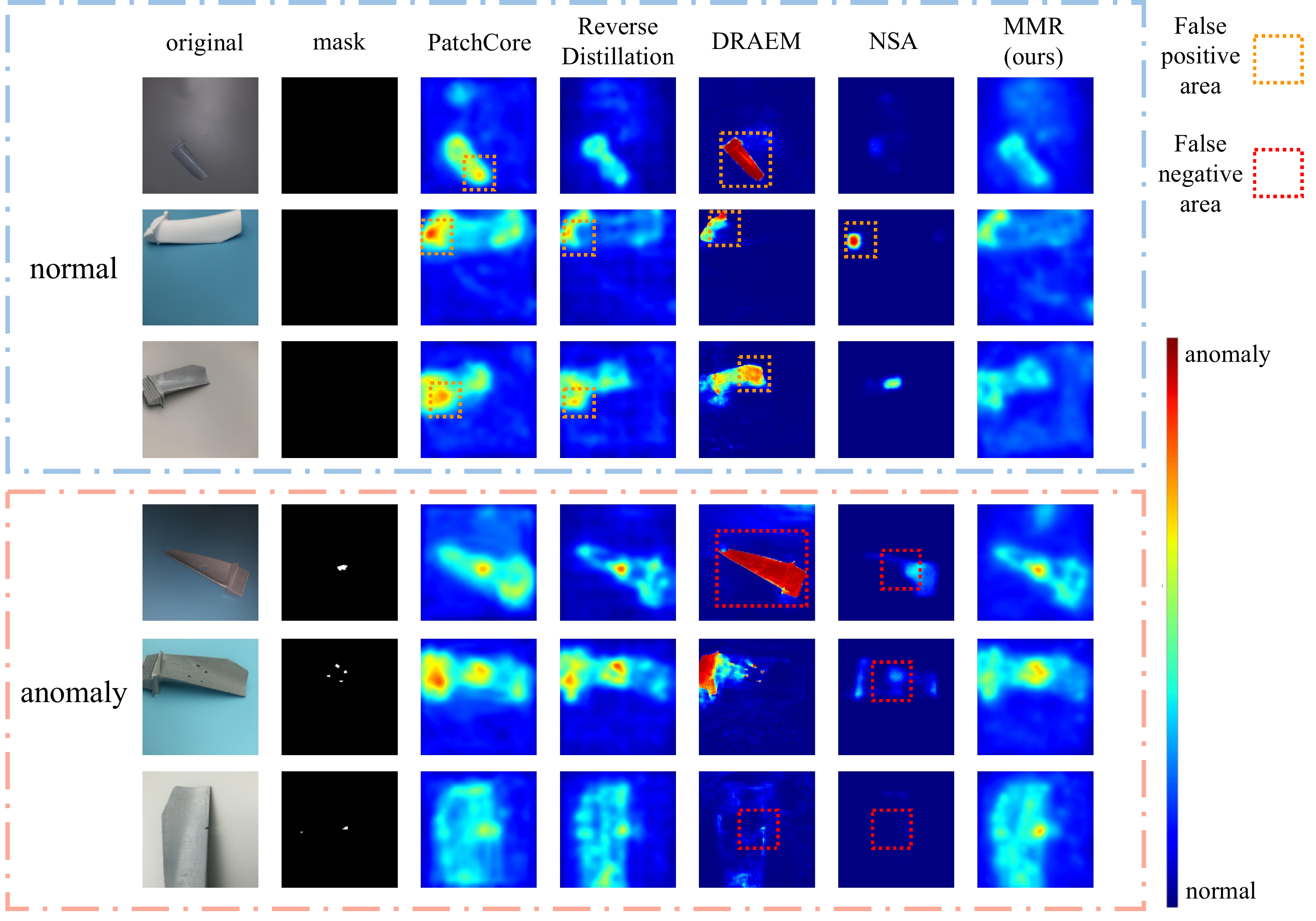}
\caption{Visualization of anomaly maps (heatmaps) of current SOTA methods and MMR. The red areas are regarded as abnormal areas, and other colors as normal areas. As the domain shifts, PatchCore and RecerseDistillation incorrectly evaluate normal regions as abnormal regions. Since the anomaly score of an image is the maximum value of the heatmap generated by that image, this makes PatchCore and RecerseDistillation produce some false positives. Conversely, DRAEM and NSA miss smaller defects to produce some false negative areas. For our proposed MMR, it can detect anomalies while reducing false positives. (\textbf{best view in color})}
\label{fig_5}
\end{figure*}

\subsection{Description of The Dataset}

To better serve the needs of blade anomaly detection in real-world scenarios and bridge the gap between current datasets and practical applications, we introduce the \textbf{A}ero-\textbf{e}ngine \textbf{B}lade \textbf{A}nomaly \textbf{D}etection (AeBAD) dataset, consisting of two sub-datasets: AeBAD-S and AeBAD-V. AeBAD-S contains images of single blades, while AeBAD-V consists of videos of multiple blades on blisks of aero-engines. The training set of AeBAD only contains normal samples, while the test sets include both normal and abnormal samples. The anomalies of the blade consist of the following four types of defects: breakdown, ablation, groove, and fracture. Breakdown is caused by foreign objects (e.g., sand, metal, birds, hail, etc.), which usually cause the blade to break \cite{yang2022review}. Ablation comes from the high temperature gas that the blades are exposed to. Ablation will reduce the performance of the blade. When the ablation reaches a certain level and the performance of the blade material cannot meet the requirements, fracture will occur. Groove is caused by a large number of elastic stress cycles, which will expand causing the fracture of the blade. Some abnormal samples are shown in Figure \ref{fig_3}. The statistical overview of AeBAD is listed in Table \ref{Table_overview}. The total training samples are 1228 and the test samples are 4342. In addition, for all anomalies present in AeBAD-S dataset, we provide pixel-level ground-truth annotations. And for all frames in AeBAD-V dataset, we provide sample-level annotations.

Different from the current datasets, where all images are roughly aligned, the shooting positions of acquired images in the training set of AeBAD-S are sampled from the position of blue dotted lines, which are enclosed by the negative half axis of x, the positive half axis of y, and the positive axis of z. The shooting positions are shown in Figure \ref{fig_2} top. Besides, there are three background colors for the training image: black, gray and blue. Some training samples are shown in the first row of Figure \ref{fig_3}. The test set of AeBAD-S is divided into four categories: same, background, illumination, and view. In the "same" category, the distribution of the normal samples in the test set is the same as that in the training set. The "background" category involves changes to the image background, while the "illumination" category involves changes to the lighting conditions. Finally, the "view" category involves changes to the shooting positions, with the new positions being symmetrical to those in the training set. The shooting positions of the “view” test set are shown in Figure \ref{fig_2} bottom. In the above test sets, there is a domain shift in all except the “same” test set. The samples of different test sets are shown in Figure \ref{fig_3}.

AeBAD-V contains videos taken from different viewpoints, and the training set consists of videos taken from four viewpoints, while the test set has videos taken from three viewpoints different from those of the training set. Some frames are shown in Figure \ref{fig_4}. The environment in AeBAD-V is more consistent with the real anomaly detection of aero-engine blades compared to AeBAD-S, which is in a simulated environment.

\begin{figure*}[!t]
\centering
\includegraphics[width=6in]{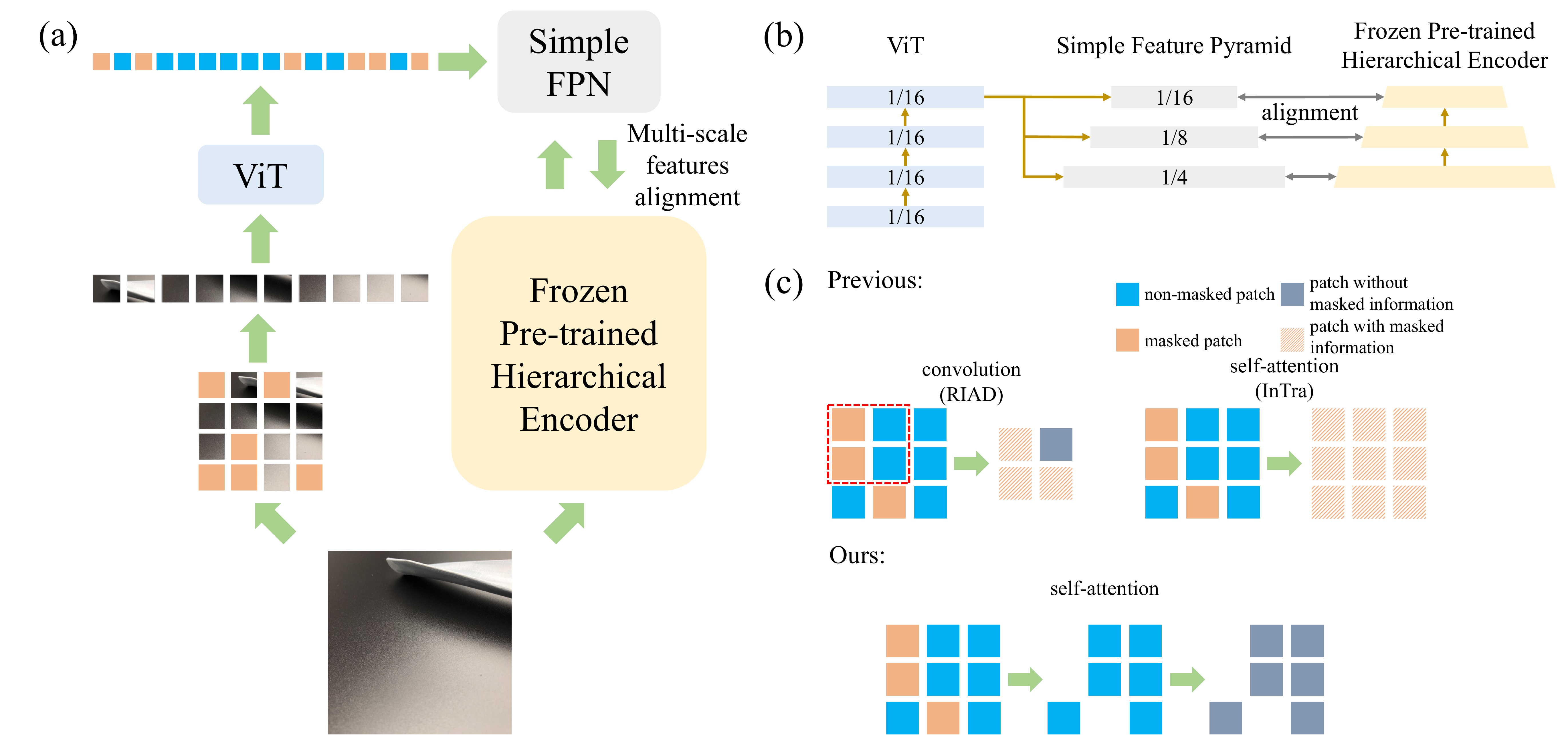}
\caption{\textbf{(a)} The overall framework of masked multi-scales reconstruction (MMR). MMR consists of two parts. \textbf{(b)} A schematic diagram of simple FPN \cite{li2022exploring}. “$\frac{1}{16}$” denotes that the receptive field of ViT's each patch is $16\times16$. \textbf{(c)} Difference between the previous methods and ours. The red dotted line indicates the convolution kernel. The previous methods share the features of masked parts with the features of the visible parts by convolution or self-attention. Our method prevents information of the visible parts leakage. (\textbf{best view in color})}
\label{fig_mmr}
\end{figure*}

\subsection{Deficiencies of The Current Methods}
\label{3_3}

The proposed AeBAD dataset enables us to evaluate the performance of current IAD methods under domain shifts. We evaluate the performance of four SOTA methods on the AeBAD-S dataset: PatchCore \cite{roth2022towards}, ReverseDistillation \cite{deng2022anomaly}, DRAEM \cite{zavrtanik2021draem}, and NSA \cite{schluter2022natural}. It should be noted that in these methods, the anomaly score of an image is the maximum value of the heatmap generated by that image.

The qualitative results of the above methods are shown in Figure \ref{fig_5}, and more results are available in Figure \ref{fig_exp_1}. In the qualitative results, the red areas are regarded as abnormal areas, and other colors as normal areas. We find the following problems: 1.) the methods based on synthetic abnormalities (e.g., NSA) tend to miss defects and produce false negative areas. For DRAEM, the cause of failure is more complicated, we will analyze it in Section \ref{5_exp}. 2.) the methods based on matching, such as PatchCore and ReverseDistillation, tend to incorrectly identify normal regions as abnormal regions under domain shifts, resulting in many false positives.

Regarding problem 1, the reason is that the synthetic anomalies are mostly distributed in background areas as the samples are not aligned and at different scales. This significantly reduces the difficulty of the task of discriminating normal samples and synthetic anomalies, resulting in poor generalization for unseen anomalies. As for problem 2, existing methods heavily rely on modeling the distribution of normal samples in the training set, which creates a discrepancy between the features of normal samples in the training and test sets when the domian shifts. We think that the problem 2 is likely due to a lack of consideration for causality in normal samples.

\begin{figure*}[!t]
\centering
\includegraphics[width=6in]{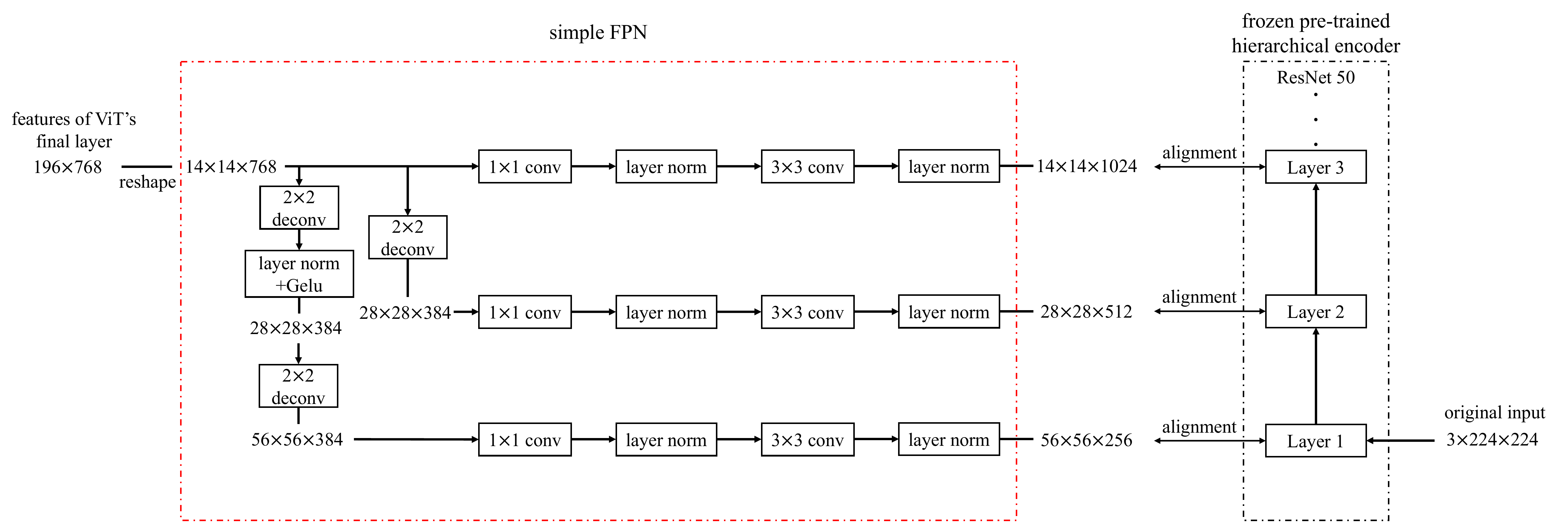}
\caption{Architecture of simple FPN \cite{li2022exploring} (\textbf{shown in the red dotted box}). The dimension of original input is assumed to be $3\times224\times224$. The patch in ViT is $16\times16$ and the dimension of ViT's feature is 768. The figure shows the case where the feature pyramid contains $\times4$, $\times2$ and $\times1$ features. The frozen pre-trained hierarchical encoder is assumed to be ResNet 50.}
\label{fig_simple_fpn}
\end{figure*}

\section{Masked Multi-scales Reconstruction}
\label{4_method}

Motivated by the above observations and analyses, we propose a synthetic anomalies free anomaly detection method entitled \textbf{M}asked \textbf{M}ulti-scales \textbf{R}econstruction (MMR). The overall framework of MMR is shown in Figure \ref{fig_mmr} (a). 

\subsection{Details of MMR}

MMR consists of two parts. In the first part, as shown in Figure \ref{fig_mmr} (a) left, the image is evenly divided into several patches (16 for Figure \ref{fig_mmr}), and a certain percentage of these patches are masked while others are preserved, where the patches have equal height and width. The preserved patches are fed into a vision transformer (ViT) \cite{dosovitskiy2020image}, which generates embeddings of these patches. The random initialized embeddings of the masked patches are then appended to the embeddings of the preserved patches. A simple feature pyramid network (simple FPN) \cite{li2022exploring} is used to take the embeddings of all patches as input and generate multi-scale features. In the second part, as shown in Figure \ref{fig_mmr} (a) right, a frozen pre-trained hierarchical encoder is used to encode the same input used in the first part into multi-scale features. The goal of MMR is to align the multi-scale features produced by the two parts. The underlying motivation for recovering the information on the masked patches is that we hope the model can perceive the dependence of spatial positions of different parts in images. In MMR, the reconstruction task is responsible for focusing on the details, while the masked recovery task is for the causality of various parts.

\noindent \textbf{Architecture of Simple FPN \cite{li2022exploring}} Simple FPN follows the idea in \cite{lin2017feature, adelson1984pyramid}. The target is to build a feature pyramid containing multi-scale features. Different from the previous work \cite{lin2017feature}, which needs to take the features of different scales as the inputs and needs lateral connection, simple FPN only needs the features of final layers of ViT to decode the features of different scales in parallel. The schematic diagram is shown in Figure \ref{fig_mmr} (b). We use the original architecture in simple FPN \cite{li2022exploring}, which is as follows: the scale which is the same as the features of final layer of ViT uses the ViT’s final feature map. Scale $\times 2$ (or $\times 4$) is built by one (or two) $2 \times 2$ deconvolution layer(s) with stride=2. In the $\times 4$ scale case, the first deconvolution is followed by LayerNorm (LN) \cite{ba2016layer} and GeLU \cite{hendrycks2016gaussian}. Then for each pyramid level, simple FPN applys a 1×1 convolution with LN and then a 3×3 convolution also with LN. The dimensions of feature map are the same as the dimensions of the features of different scales in the frozen pre-trained hierarchical encoder. The architecture of simple FPN is shown in Figure \ref{fig_simple_fpn}.

\noindent \textbf{Problem Formulation} Formally, in the first part, to handle 2D images, we firstly reshape the original image $\textbf{x} \in h\times w\times c$ into a sequence of ﬂattened 2D patches $\textbf{x}_{p} \in n\times (p^2 \times c)$, where $(h, w)$ is the resolution of the original image, $c$ is the number of channels, $n=\frac{hw}{p^2}$ is the number of patches, and $(p, p)$ is the resolution of patch. Then, we use the operation $\Phi(\cdot; \eta)$ to randomly extract $n_{masked} = \lfloor n\times (1-\eta) \rfloor$ ﬂattened patches as non-masked patches, where $\eta$ is masking ratio. After that, we use a ViT $f_{ViT}(\cdot)$ to encode the non-masked patches, where each layer of ViT consists of a multi-head self-attention module and a feed forward network. The embeddings of masked patches, which are randomly initialized, are appended to the output $f_{ViT}(\Phi(\textbf{x}_{p}; \eta)) \in n_{masked}\times (p^{2} \times c)$ of ViT by the operation $\Psi(\cdot)$. Finally, the embeddings $\Psi(f_{ViT}(\Phi(\textbf{x}_{p}; \eta))) \in n\times (p^{2} \times c)$ of all patches are reshaped into $\sqrt{n} \times \sqrt{n} \times (p^{2} \times c)$, where the receptive field of every feature on the feature map is $(p, p)$. The simple FPN $f_{FPN_{i}}(\cdot)$ decodes all embeddings to the features $\textbf{z}_{masked_{i}}$ of different scales, where $i$ denotes the i-th scales. In the second part, a frozen pre-trained hierarchical encoder $f_{FPH}(\cdot)$ encodes $\textbf{x}$ into the features $\textbf{z}_{frozen_{i}} \in (h_{i} \times w_{i}) \times c_{i}$ of different scales, where $h_{i}, w_{i}, c_{i}$ are height, width and channels respectively. And the dimensions of $\textbf{z}_{masked_{i}}$ are the same as $\textbf{z}_{frozen_{i}}$. Note that $h_{i}$ and $w_{i}$ are $2^{i-1} \times \sqrt{n}$. The final loss function of MMR is as follows:

\begin{equation}
\label{EN_1}
\mathcal L = \sum\limits_{i=1}\limits^{s} \sum\limits_{k=1}\limits^{h_{i} \times w_{i}} \frac{1}{h_{i} \times w_{i}}(1 - \frac{\textbf{z}_{masked_{i}}(k) {(\textbf{z}_{frozen_{i}}(k))}^T}{||\textbf{z}_{masked_{i}}(k)|| ||\textbf{z}_{frozen_{i}}(k)||}),
\end{equation}

\noindent where $s$ denotes the total scales and $\textbf{z}_{masked_{i}}(k) \in 1\times c_{i}$ denotes the k-th row of $\textbf{z}_{masked_{i}}$. Our target is to minimize the Eq. (\ref{EN_1}) to recover the multi-scale features of masked patches.

In the test stage, MMR takes the original \textbf{non-masked} input (overall image) to the above two parts. The first part and the second part generate the features $\textbf{z}_{masked_{i}}$ and $\textbf{z}_{frozen_{i}}$ respectively. The anomaly map $\rm{AM}_{i} \in h_{i} \times w_{i}$ of every scale $i$ is calculated as follows:

\begin{equation}
\label{EN_2}
\rm{AM}_{i} = \zeta((\textbf{z}^{\prime}_{masked_{i}} \odot \textbf{z}^{\prime}_{frozen_{i}}) \mathbbm{1}), 
\end{equation}

\noindent where $\textbf{z}^{\prime}_{masked_{i}}$ is the result of $\textbf{z}_{masked_{i}}$ normalized by row, $\odot$ is hadamard product, $\mathbbm{1} \in c_{i} \times 1$ is a vector whose elements are all 1 and $\zeta(\cdot)$ is an operation which reshapes the vector $v \in (h_{i} \times w_{i}) \times 1$ to $v^{\prime} \in h_{i} \times w_{i}$. For the feature maps with different resolutions, they are uniformly up-sampled to the resolution of original image by bilinear interpolation. \textbf{Anomaly map:} The final anomaly map $\rm{AM}$ is the sum of up-sampled $\rm{AM}_{i}$. \textbf{Anomaly score:} The anomaly score of one image is the largest value on $\rm{AM}$.

\noindent \textbf{Mechanism of MMR} During the training stage, MMR uses normal samples as input and recovers multi-scale features of masked patches based on the information provided by non-masked patches. This enables ViT and simple FPN to capture causal relationships among different parts of normal samples and avoid learning an identical function that is not helpful for anomaly detection, where the causal relationships among different parts mean the dependence of spatial positions of various parts. Specifically, the human can predict the overall position of the blade and some details on the blade by only capturing the part of the root of the blade. In MMR, this process is replaced by predicting the masked patches by the non-masked patches. Mathematically, this process is equivalent to modeling the \textcolor{red}{conditional distribution}, where the non-masked part is the condition and the masked part is the target. As mentioned in Section \ref{3_3}, PatchCore and Reverse Distillation aim to model the distribution of the normal samples (unconditional distribution), which creates a discrepancy between the features of normal samples in the training and test sets when the domian shifts. Unlike these two methods, MMR models the conditional distribution among different parts. When the domain shifts, the conditional distribution among image patches remains relatively unchanged despite the change in the distribution of the image. This makes MMR relatively insensitive to domain shifts. 

During testing, the original non-masked input is used for both parts of the method. Since the frozen pre-trained hierarchical encoder contains the rich semantics, it can perceive the unseen anomalies. And for ViT and simple FPN, since they only model causality in normal sample, it cannot reconstruct the abnormal parts well. This leads to biases in the reconstructed and encoded multi-scale features for anomalies, where the biases correspond to the abnormal parts in the feature map.

\begin{figure}[!t]
\centering
\includegraphics[width=3in]{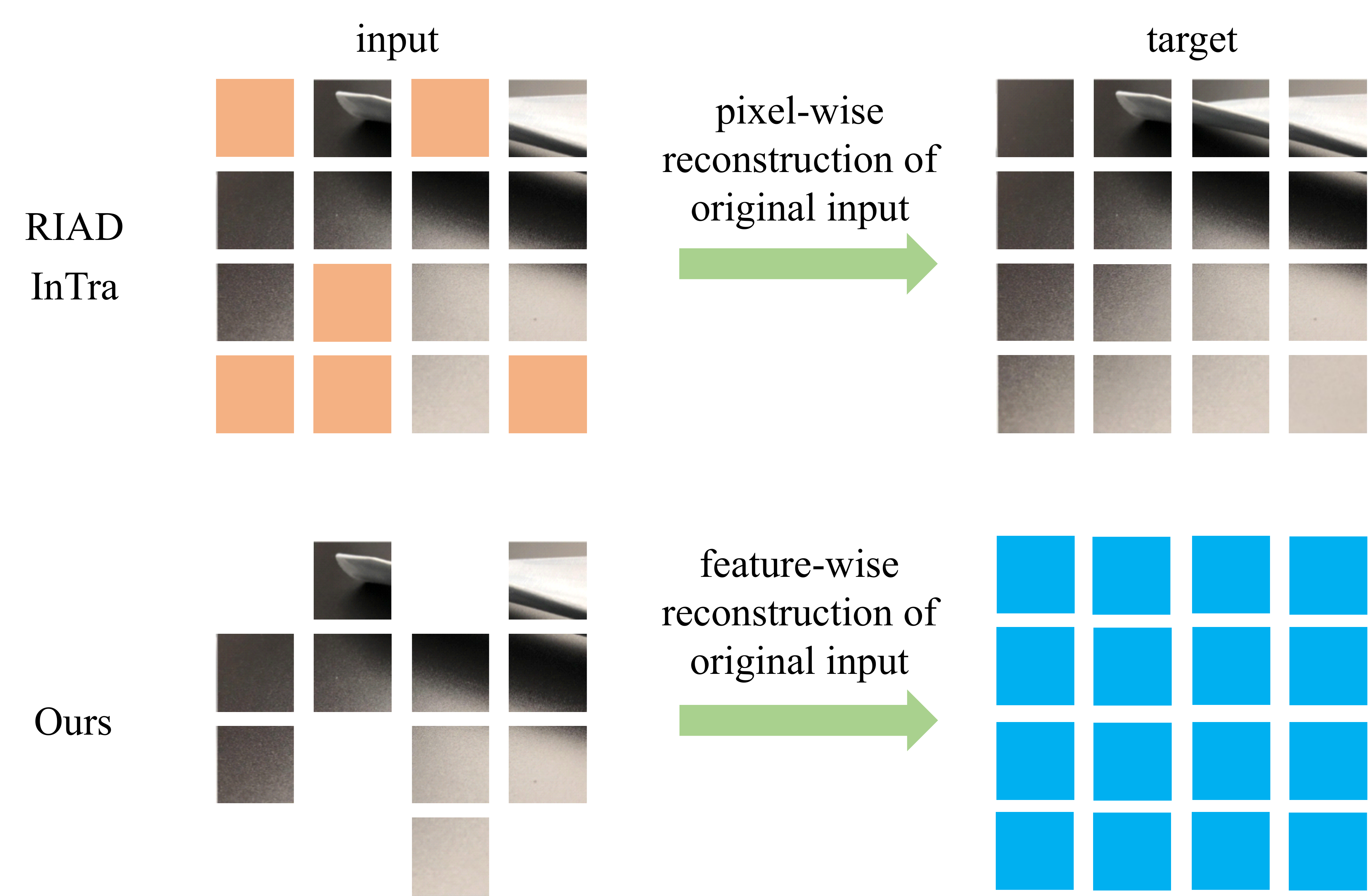}
\caption{Difference of reconstructed target between the previous methods and ours.}
\label{fig_dif_res}
\end{figure}

\begin{table*}
    \scriptsize
    \centering
    \caption{Sample-level anomaly detection performance (AUROC \%) on AeBAD-S dataset. Bold denotes the best results. The results over 5 trials are reported.}
    \begin{tabular}{c c c c c c}
    \hline
    Method & Same & Background & Illumination & View & Mean \\ \hline
    PatchCore \cite{roth2022towards} {\fontsize{5pt}{2pt}\selectfont CVPR' 2022} & 75.2 $\pm$ 0.3 & 74.1 $\pm$ 0.3 & 74.6 $\pm$ 0.4 & 60.1 $\pm$ 0.4 &  71.0 \\
    ReverseDistillation \cite{deng2022anomaly} {\fontsize{5pt}{2pt}\selectfont CVPR' 2022} & 82.4 $\pm$ 0.6 & 84.3 $\pm$ 0.9 & 85.5 $\pm$ 0.9 & 71.9 $\pm$ 0.8 & 81.0 \\
   DRAEM \cite{zavrtanik2021draem} {\fontsize{5pt}{2pt}\selectfont ICCV' 2021} & 64.0 $\pm$ 0.4 & 62.1 $\pm$ 6.1 & 61.6 $\pm$ 2.7 & 62.3 $\pm$ 0.9 & 62.5 \\ 
   NSA \cite{schluter2022natural} {\fontsize{5pt}{2pt}\selectfont ECCV' 2022} & 66.5 $\pm$ 1.4 & 48.8 $\pm$ 3.5 & 55.5 $\pm$ 3.2 & 55.9 $\pm$ 1.1 & 56.7 \\ 
   RIAD \cite{zavrtanik2021reconstruction} {\fontsize{5pt}{2pt}\selectfont PR' 2020} & 38.6 $\pm$ 0.6 & 41.6 $\pm$ 1.3 & 46.8 $\pm$ 0.8 & 33.0 $\pm$ 0.6 & 40.0\\ 
   InTra \cite{pirnay2022inpainting} {\fontsize{5pt}{2pt}\selectfont ICIAP' 2022} & 39.8 $\pm$ 0.8 & 46.1 $\pm$ 0.5 & 44.7 $\pm$ 0.3 & 46.3 $\pm$ 1.5 & 44.2\\ \hline
   MMR (\textbf{Ours}) & \textbf{85.6} $\pm$ 0.5 & \textbf{84.4} $\pm$ 0.7 & \textbf{88.8} $\pm$ 0.5 & \textbf{79.9} $\pm$ 0.6 & \textbf{84.7} \\ \hline
    \end{tabular}
    \label{Table_1}
\end{table*}

\begin{table*}
    \scriptsize
    \centering
    \caption{Pixel-level anomaly detection performance (PRO \%) on AeBAD-S dataset. Bold denotes the best results. The results over 5 trials are reported.}
    \begin{tabular}{c c c c c c}
    \hline
    Method & Same (PRO) & Background (PRO) & Illumination (PRO) & View (PRO) & Mean (PRO)\\ \hline
    PatchCore \cite{roth2022towards} {\fontsize{5pt}{2pt}\selectfont CVPR' 2022} & 89.5 $\pm$ 0.2 & 89.4 $\pm$ 0.1 & 88.2 $\pm$ 0.1 & 84.0 $\pm$ 0.2 & 87.8 \\
    ReverseDistillation \cite{deng2022anomaly} {\fontsize{5pt}{2pt}\selectfont CVPR' 2022} & 86.4 $\pm$ 0.4 & 86.4 $\pm$ 0.7 & 86.7 $\pm$ 0.5 & 82.9 $\pm$ 0.7 & 85.6 \\
   DRAEM \cite{zavrtanik2021draem} {\fontsize{5pt}{2pt}\selectfont ICCV' 2021} & 71.4 $\pm$ 4.2 & 44.3 $\pm$ 11.6 & 67.6 $\pm$ 2.7 & 71.1 $\pm$ 2.3 & 63.6 \\ 
   NSA \cite{schluter2022natural} {\fontsize{5pt}{2pt}\selectfont ECCV' 2022} & 43.0 $\pm$ 1.3 & 29.7 $\pm$ 2.1 & 59.9 $\pm$ 1.3 & 51.1 $\pm$ 0.1 & 45.9 \\
   RIAD \cite{zavrtanik2021reconstruction} {\fontsize{5pt}{2pt}\selectfont PR' 2020} & 71.9 $\pm$ 1.3 & 33.4 $\pm$ 0.6 & 65.3 $\pm$ 1.0 & 62.2 $\pm$ 1.7 & 58.2\\ 
   InTra \cite{pirnay2022inpainting} {\fontsize{5pt}{2pt}\selectfont ICIAP' 2022} & 76.8 $\pm$ 0.2 & 74.8 $\pm$ 0.3 & 73.7 $\pm$ 0.3 & 73.4 $\pm$ 0.2 & 74.7\\ \hline
   MMR (\textbf{Ours}) & \textbf{89.6} $\pm$ 0.2 & \textbf{90.1} $\pm$ 0.2 & \textbf{90.2} $\pm$ 0.2 & \textbf{86.3} $\pm$ 0.3 & \textbf{89.1} \\ \hline
    \end{tabular}
    \label{Table_2}
\end{table*}

\subsection{\textbf{Relations between MMR and Current Methods}}

\noindent \textbf{Relations to the inpainting methods} There are two main ways in which MMR differs from current methods based on reconstruction. First, as shown in Figure \ref{fig_dif_res}, MMR does not try to reconstruct all the details of the full image, but instead focuses on recovering features at different scales. Since the frozen encoder is insensitive to some changes  \cite{bhardwaj2023steerable}, the encoded features are more robust. This allows MMR to emphasize causal relationships among different parts of normal samples. Second, MMR uses a different masked strategy compared to previous methods like RIAD and InTra, as shown in Figure \ref{fig_mmr} (c). In MMR, the masked parts are completely hidden, while in RIAD \cite{zavrtanik2021reconstruction} (convolution in Figure \ref{fig_mmr} (c)) and InTra \cite{pirnay2022inpainting} (ViT in Figure \ref{fig_mmr} (c)), the features of the masked parts are shared with the features of the visible parts. MMR's approach prevents information from the visible parts from leaking into the masked parts. This can be further verified by the ablation study in Section \ref{6_abl}.

\noindent \textbf{Relations to the student-teacher framework} The student-teacher framework, also known as knowledge distillation \cite{hinton2015distilling}, is an efficient method for handling the one-class anomaly detection task. Current methods, e.g., MKDAD \cite{salehi2021multiresolution} and ReverseKD \cite{deng2022anomaly}, distilled the pre-trained features for the normal samples. MMR can also be regarded as a student-teacher network, where the frozen pre-trained encoder plays the role of the teacher and another part is a student network. The unique difference between MMR and current knowledge distillation methods is that there is a causal inference module in the knowledge distillation. This endows the model with the perception of spatial positions of different parts in images so that it can improve the generalization in the scenario under the domain shift.

\noindent \textbf{Relations to masked autoencoder (MAE) \cite{he2022masked}} MAE adopted a same masked strategy as MMR. Although the masked strategy is the same, the intention of the masked strategy is entirely different. MAE used such a masked strategy, i.e., only training on a small subset of image patches, to reduces the consumption of memory so that the larger batch sizes can be allowed, while MMR uses this strategy to help the model perceive the dependence of spatial positions of different parts in images and model the causality in normal sample. More importantly, we provide a new insight into such a masked strategy, where it can prevent information from the visible parts from leaking into the masked parts (as described in Figure \ref{fig_mmr} (c)). Another difference between MMR and MAE is the reconstruction target, where MAE reconstructed the original pixel-wise input (similar to the top of Figure \ref{fig_dif_res}), while MMR performs a feature-wise reconstruction.

\subsection{Implementation Details}
\label{sec_id}

The input image, augmented with random cropping, is resized to $256 \times 256$ and then center croped to $224 \times 224$. The resolution of patch is (16, 16). The masking ratio $\eta$ is 0.4. The backbone of ViT is the vanilla ViT-B \cite{dosovitskiy2020image}, where the backbone is pre-trained by masked autoencoder \cite{he2022masked}. The simple FPN generates feature maps of 4, 8, and 16 receptive fields. We use a widely adopted WideResNet50 \cite{zagoruyko2016wide, he2016deep} as the frozen pre-trained hierarchical encoder and then extract the multi-scale features from the layer 1, 2 and 3. We use AdamW optimizer \cite{loshchilov2017decoupled} ($\beta_{1}=0.9, \beta_{2}=0.95$) with step-wise learning rate decay. The learning rate is 0.001. For all datasets, we set training epoch as 200 and batch size as 16. In addition, all experiments were conducted on Ubuntu 18.04.5 and a computer equipped with Xeon(R) Gold 6140R CPUs@2.30GHZ and an NVIDIA GeForce RTX 2080 Ti with 11GB of memory.

\section{Experiments}
\label{5_exp}

\subsection{Datasets and Comparison Methods}

We evaluate MMR on AeBAD dataset including AeBAD-S and AeBAD-V. In addition, we also evaluate MMR on MVTec dataset to show the ability to detect different types of anomalies. The comparison methods include the recent SOTA methods and the methods most related to our work, including PatchCore \cite{roth2022towards}, ReverseDistillation \cite{deng2022anomaly}, DRAEM \cite{zavrtanik2021draem}, NSA \cite{schluter2022natural}, RIAD \cite{zavrtanik2021reconstruction} and InTra \cite{pirnay2022inpainting}. For PatchCore, ReverseDistillation, DRAEM and NSA, we use the official codes, where the backbone of PatchCore and ReverseDistillation is a pre-trained WideResNet50. We tune the hyperparameters to make sure the optimal performance. For RIAD and InTra, since they do not provide the official code, we use the unofficial re-implementations of these methods \footnote{The unofficial re-implementations of RIAD and InTra are \href{https://github.com/taikiinoue45/RIAD}{here} and \href{https://github.com/jhy12/inpainting-transformer}{here} respectively.}.

\begin{figure*}[!t]
\centering
\includegraphics[width=5in]{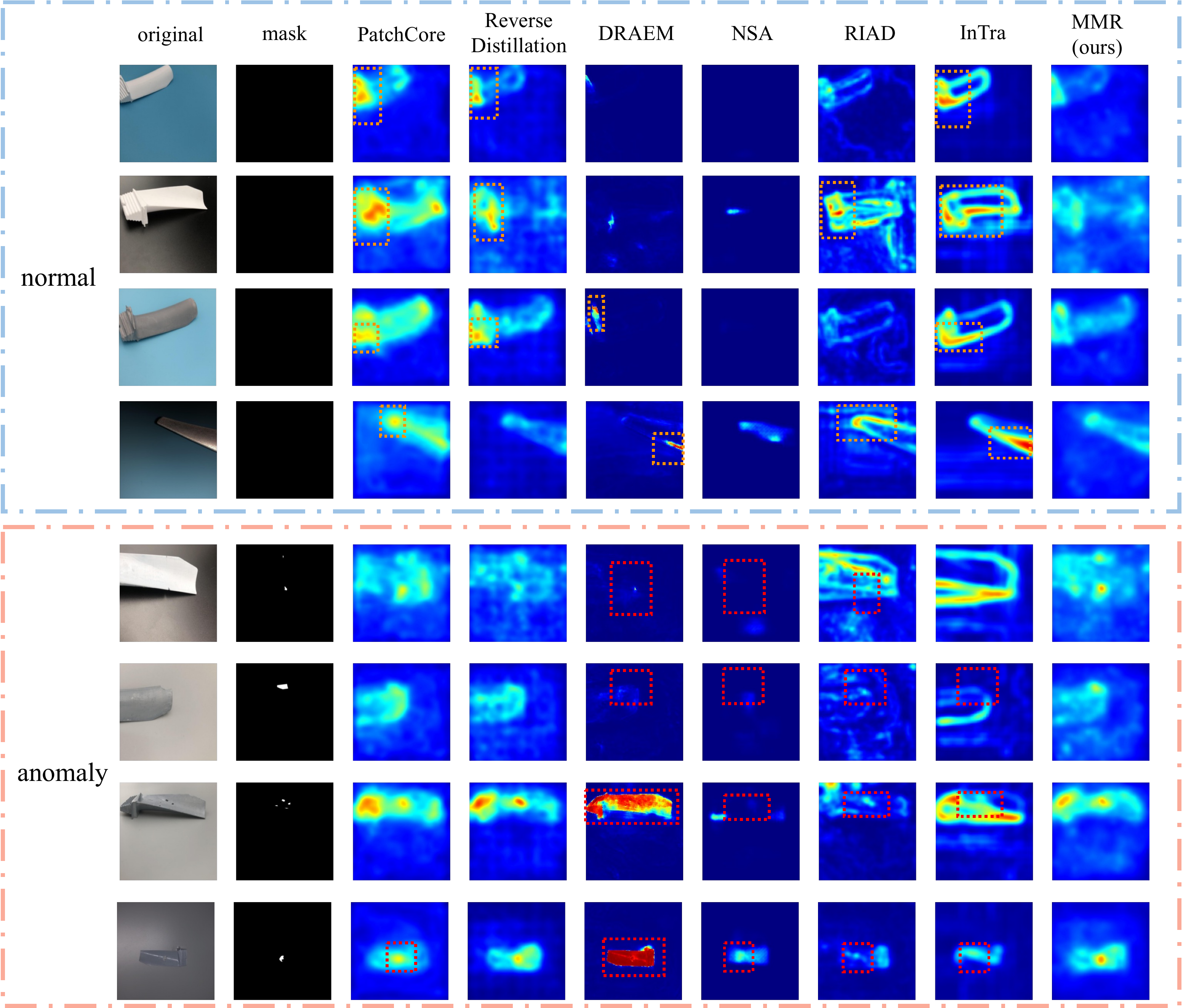}
\caption{Visualization of anomaly maps (heatmaps) of current SOTA methods, the most related works and MMR on AeBAD-S dataset. The meaning of the dotted box and heatmap are the same as those in Figure \ref{fig_5}. (\textbf{best view in color})}
\label{fig_exp_1}
\end{figure*}

\begin{figure*}[!t]
\centering
\includegraphics[width=6in]{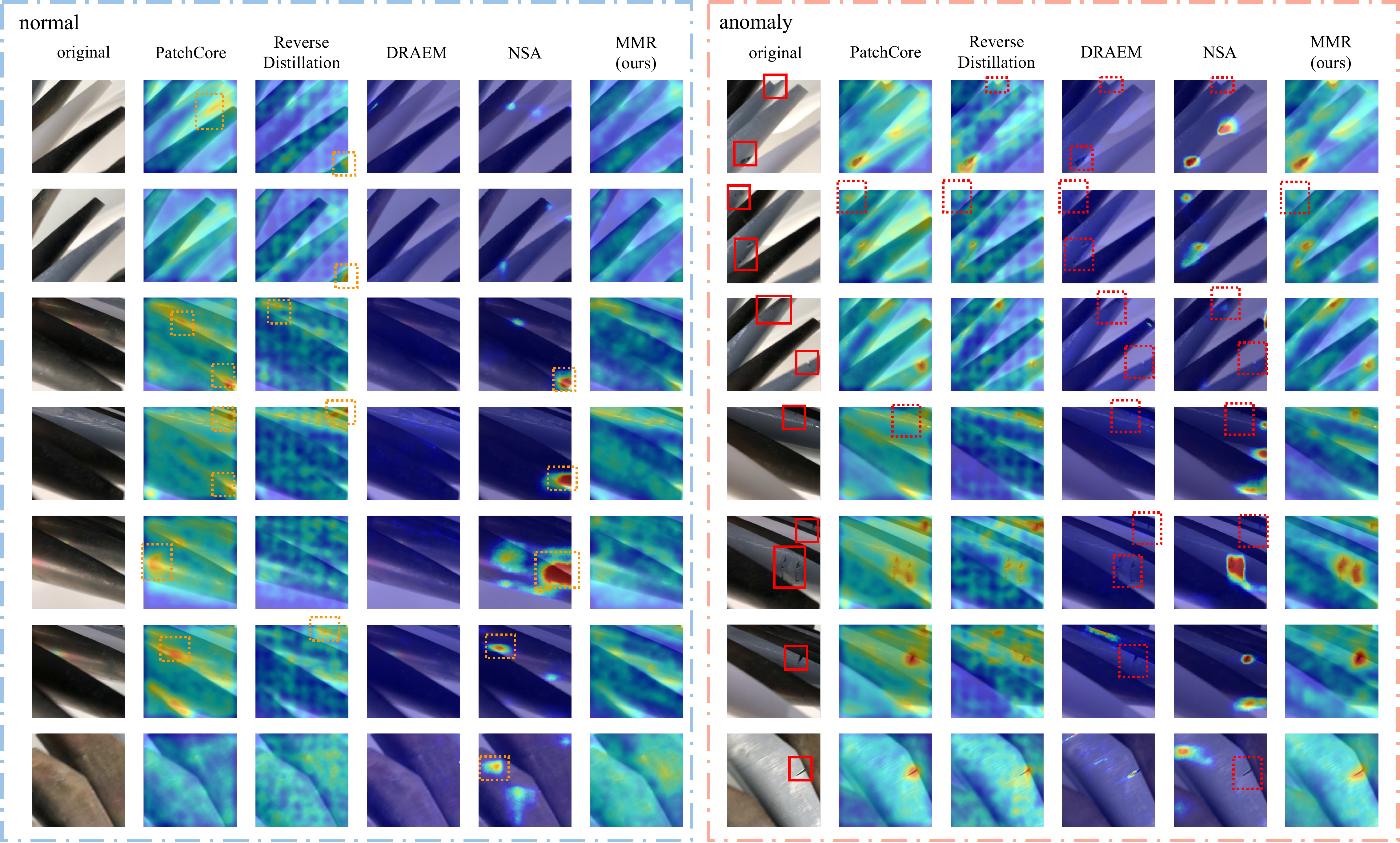}
\caption{Visualization of anomaly maps (heatmaps) of current SOTA methods, the most related works and MMR on AeBAD-V dataset. The meaning of the dotted box and heatmap are the same as those in Figure \ref{fig_5}. The red solid line box indicates the bounding box of the abnormal area. (\textbf{best view in color})}
\label{fig_exp_2}
\end{figure*}

\subsection{Results}

\noindent \textbf{AeBAD-S:} Metric: To evaluate sample-level anomaly detection, we use the area under the receiver operating characteristic (AUROC), which is a widely used metric that takes into account both false positives and false negatives. For pixel-level anomaly detection, we use the per-region-overlap (PRO) score \cite{bergmann2020uninformed} as the evaluation metric. The PRO score treats anomalies of any size equally. Unlike the common pixel-level AUROC, the PRO score decreases significantly when disconnected domains of anomalies are not detected.

The quantitative results are shown in Table \ref{Table_1} and \ref{Table_2}. The qualitative results are shown in Figure \ref{fig_exp_1}. We can observe the following phenomena: 1.) Except for DRAEM, all methods show a significant drop in performance on the "view" subset compared to other subsets. This is mainly because changes in view increase false positives for normal samples, as shown in Figure \ref{fig_5} and \ref{fig_exp_1}. In the third row of abnormal samples of Figure \ref{fig_exp_1}, MMR also produces a high score for the normal region. 2.) The methods based on the pre-trained model, such as PatchCore, ReverseDistillation and MMR, are robust to the change for background and illumination. In addition, compared to PatchCore and ReverseDistillation, our proposed method can reduce the anomaly score of the normal region as the domain shifts, as shown in the normal part of Figure \ref{fig_exp_1}. Note that the anomaly score of an image is the maximum value of the heatmap generated by that image. PatchCore and ReverseDistillation produce more false positive areas on the "view" datasets, which is also the reason their AUROC performance on the "view" dataset is poor, although their PRO performance is close to MMR's. 3.) NSA show a poor performance for anomaly classification and localization and show a large variance compared to other methods. This is mainly because the proposed dataset is not aligned, and the augmented synthetic anomalies can significantly affect performance. When the synthetic anomalies are distributed less on the background region, the performance is good, and vice versa. 4.) For DRAEM, since it generates the synthetic anomalies distributed at the full image, it can detect some apparent anomalies (e.g., the second row of the anomaly part in Figure \ref{fig_5}). However, when the light and background change, it regards the full blade as an anomaly (e.g., the first and fourth row of abnormal samples in Figure \ref{fig_5} and \ref{fig_exp_1}). We think that the main reason for this is that DREAM overfits the data due to its two large models, which results in a poor generalization for data with domain shift. 5.) RIAD and InTra use pixel-level reconstruction, which leads to larger reconstruction errors for complex details.

In addition, although the input of the first part of MMR is a whole image in the test stage, the output of this part is not identical with the output of the frozen pre-trained hierarchical encoder. This can be observed from Figure \ref{fig_5} and Figure \ref{fig_exp_1}, where MMR can detect some tiny defects (anomalies). This result demonstrates that the first part of MMR can avoid an “identical shortcut”.

\begin{table}
    \scriptsize
    \centering
    \caption{Sample-level anomaly detection performance (AUROC \%) on AeBAD-V dataset. Bold denotes the best results. The results over 5 trials are reported.}
    \begin{tabular}{c c c c c}
    \hline
    Method & Video 1 & Video 2 & Video 3 & Mean \\ \hline
    PatchCore \cite{roth2022towards} & 71.1 $\pm$ 0.2 & 86.0 $\pm$ 0.4 & 55.1 $\pm$ 0.7 & 70.7 \\
    ReverseDistillation \cite{deng2022anomaly} & 66.0 $\pm$ 2.1 & 84.8 $\pm$ 1.8 & 62.1 $\pm$ 1.1 & 71.0 \\
   DRAEM \cite{zavrtanik2021draem} & \textbf{79.5} $\pm$ 5.0 & 71.2 $\pm$ 3.3 & 53.6 $\pm$ 1.2 & 68.1 \\ 
   NSA \cite{schluter2022natural} & 59.4 $\pm$ 3.5 & 72.7 $\pm$ 1.3 & 61.9 $\pm$ 4.0 & 64.6 \\ 
   RIAD \cite{zavrtanik2021reconstruction} & 78.0 $\pm$ 1.2 & 47.1 $\pm$ 2.9 & 43.2 $\pm$ 1.1 & 56.1\\ 
   InTra \cite{pirnay2022inpainting} & 62.7 $\pm$ 1.4 & 55.8 $\pm$ 1.3 & 43.7 $\pm$ 1.4 & 54.1 \\ \hline
   MMR (\textbf{Ours}) & 75.7 $\pm$ 0.2 & \textbf{88.3} $\pm$ 1.4 & \textbf{70.7} $\pm$ 0.6 & \textbf{78.2} \\ \hline
    \end{tabular}
    \label{Table_3}
\end{table}

\noindent \textbf{AeBAD-V:} Metric: We only evaluate sample-level anomaly detection by AUROC.

The quantitative results are shown in Table \ref{Table_3}. The qualitative results are shown in Figure \ref{fig_exp_2}. Similar to the previous results, we observe that when the viewpoint changes, particularly for Video 2 and Video 3, PatchCore and ReverseDistillation produce more false positives, whereas DRAEM and NSA produce more false negatives. Our proposed method, MMR, is able to reduce the anomaly score of the normal region and detect multiple abnormal regions. It is worth noting that compared to AeBAD-S, the objects in AeBAD-V are more distributed in the center area. This allows NSA to locate apparent anomalies in some cases, but DRAEM still fails to predict anomalies. We believe that the main reason for this is also from the overfitting in DREAM. For the result of DRAEM on Video 1, since there is a similar view between video 1 and the training set, DRAEM is able to obtain a good result. In addition to the above results, we also present the qualitative results of the consecutive frames (videos) at \href{https://github.com/zhangzilongc/MMR}{https://github.com/zhangzilongc/MMR}.

\noindent \textbf{MVTec:} Metric: We follow the common protocol. For the sample-level anomaly detection, we use AUROC. For the pixel-level anomaly detection, we use PRO and pixel-level AUROC. In addition, we follow the common setups to train and test on each product category.

\begin{table*}
\scriptsize
    \centering
    \begin{tabular}{c c c c c c c c}
    \hline
    & PatchCore \cite{roth2022towards} & RD \cite{deng2022anomaly} & DRAEM \cite{zavrtanik2021draem} & NSA \cite{schluter2022natural}& RIAD \cite{zavrtanik2021reconstruction} & InTra \cite{pirnay2022inpainting} & MMR (\textbf{Ours})\\ \hline
    Bottle & 100 & 100 & 99.2 & 97.7 & 99.9 & 100 & 100 \\ 
    Cable  & 99.5 & 95.0 & 91.8 & 94.5 & 81.9 & 84.2 & 97.8 \\
    Capsule & 98.1 & 96.3 & 98.5 & 95.2 & 88.4 & 86.5 & 96.9 \\
    Carpet  & 98.7 & 98.9 & 97.0 & 95.6 & 84.2 & 98.8 & 99.6 \\
    Grid  & 98.2 & 100 & 99.9 & 99.9 & 99.6 & 100 & 100 \\
    Hazelnut & 100 & 99.9 & 100 & 94.7 & 83.3 & 95.7 & 100 \\
    Leather & 100 & 100 & 100 & 99.9 & 100 & 100 & 100 \\
    Metal Nut & 100 & 100 & 98.7 & 98.7 & 88.5 & 96.9 & 99.9 \\
    Pill  & 96.6 & 96.6 & 98.9 & 99.2 & 83.8 & 90.2 & 98.2 \\ 
    Screw & 98.1 & 97.0 & 93.9 & 90.2 & 84.5 & 95.7 & 92.5\\ 
    Tile & 98.7 & 99.3 & 99.6 & 100 & 98.7 & 98.2 & 98.7 \\
    Toothbrush & 100 & 99.5 & 100 & 100 & 100 & 99.7 & 100 \\ 
    Transistor & 100 & 96.7 & 93.1 & 95.1 & 90.9 & 95.8 & 95.1 \\  
    Wood & 99.2 & 99.2 & 99.1 & 97.5 & 93.0 & 98.0 & 99.1 \\  
    Zipper & 99.4 & 98.5 & 100 & 99.8 & 98.1 & 99.4 & 97.6 \\  \hline
     Mean & \textbf{99.1} & 98.5 & 98.0 & 97.2 & 91.7 & 95.9 & 98.4 \\\hline
    \end{tabular}
     \caption{Sample-level anomaly detection performance (AUROC \%) on MVTec.}
    \label{mvtec_1}
\end{table*}

\begin{table*}
\scriptsize
    \centering
    \begin{tabular}{c c c c c c c c}
    \hline
    & PatchCore \cite{roth2022towards} & RD \cite{deng2022anomaly} & DRAEM \cite{zavrtanik2021draem} & NSA \cite{schluter2022natural}& RIAD \cite{zavrtanik2021reconstruction} & InTra \cite{pirnay2022inpainting} & MMR (\textbf{Ours})\\ \hline
    Bottle & 98.6/96.1 & 98.7/96.6 & 99.1/- & 98.3/- & 98.4/- & 97.1/- & 98.3/96.0 \\ 
    Cable  & 98.5/92.6 & 97.4/91.0 & 94.7/- & 96.0/- & 84.2/- & 93.2/- & 95.4/87.2\\
    Capsule & 98.9/95.5 & 98.7/95.8 & 94.3/- & 97.6/- & 92.8/- & 97.7/- & 98.0/94.5 \\
    Carpet  & 99.1/96.6 & 98.9/97.0 & 95.5/- & 95.5/- & 96.3/- & 99.2/- & 98.8/96.6 \\
    Grid  & 98.7/95.9 & 99.3/97.6 & 99.7/- & 99.2/- & 98.8/- & 99.4/- & 99.0/96.5 \\
    Hazelnut & 98.7/93.9 & 98.9/95.5 & 99.7/- & 97.6/- & 96.1/- & 98.3/- & 98.5/91.2 \\
    Leather & 99.3/98.9 & 99.4/99.1 & 98.6/- & 99.5/- & 99.4/- & 99.5/- & 99.2/98.6 \\
    Metal Nut & 98.4/91.3 & 97.3/92.3 & 99.5/- & 98.4/- & 92.5/- & 93.3/- & 95.9/88.6 \\
    Pill  & 97.6/94.1 & 98.2/96.4 & 97.6/- & 98.5/- & 95.7/- & 98.3/- & 98.4/96.1\\ 
    Screw & 99.4/97.9 & 99.6/98.2 & 97.6/- & 96.5/- & 89.1/- & 99.5/- & 99.5/97.6 \\ 
    Tile & 95.9/87.4 & 95.6/90.6 & 99.2/- & 99.3/- & 89.1/- & 94.4/- & 95.6/90.2 \\
    Toothbrush & 98.7/91.4 & 99.1/94.5 & 98.1/- & 94.9/- & 98.9/- & 99.0/- & 98.4/93.0\\ 
    Transistor & 96.4/83.5 & 92.5/78.0 & 90.9/- & 88.0/- & 87.7/- & 96.1/- & 90.2/79.1 \\  
    Wood & 95.1/89.6 & 95.3/90.9 & 96.4/- & 90.7/- & 85.8/- & 90.5/- & 94.8/88.9 \\  
    Zipper & 98.9/97.1 & 98.2/95.4 & 98.8/- & 94.2/- & 97.8/- & 99.2/- & 98.0/95.0 \\  \hline
     Mean & \textbf{98.1}/93.5 & 97.8/\textbf{93.9} & 97.3/- & 96.3/- & 94.2/- & 97.0/- & 97.2/92.6 \\\hline
    \end{tabular}
     \caption{Pixel-level anomaly detection performance (pixel-level AUROC \% / PRO) on MVTec.}
    \label{mvtec_2}
\end{table*}

The quantitative results are shown in Table \ref{mvtec_1} and \ref{mvtec_2}. We can observe that MMR also obtain the competitive results with SOTA methods. It demonstrates the detection ability of MMR for the anomalies of different types. It is worth noting that compared with the superior performance in AeBAD dataset, the performance of MMR in MVTec dataset is suboptimal. The main reason is that when the distribution of the test sample changes, MMR can better judge normal samples and abnormal samples, while other methods will be affected by domain shift. However, for the MVTec dataset, the distribution of test samples is consistent with the training set, so the advantages of MMR cannot be highlighted. In addition, we observe that compared with PatchCore, the acquired abnormal masks of MMR are coarser, which is also the reason for the suboptimal performance in the pixel-level performance.

\begin{figure}[!t]
\centering
\includegraphics[width=3in]{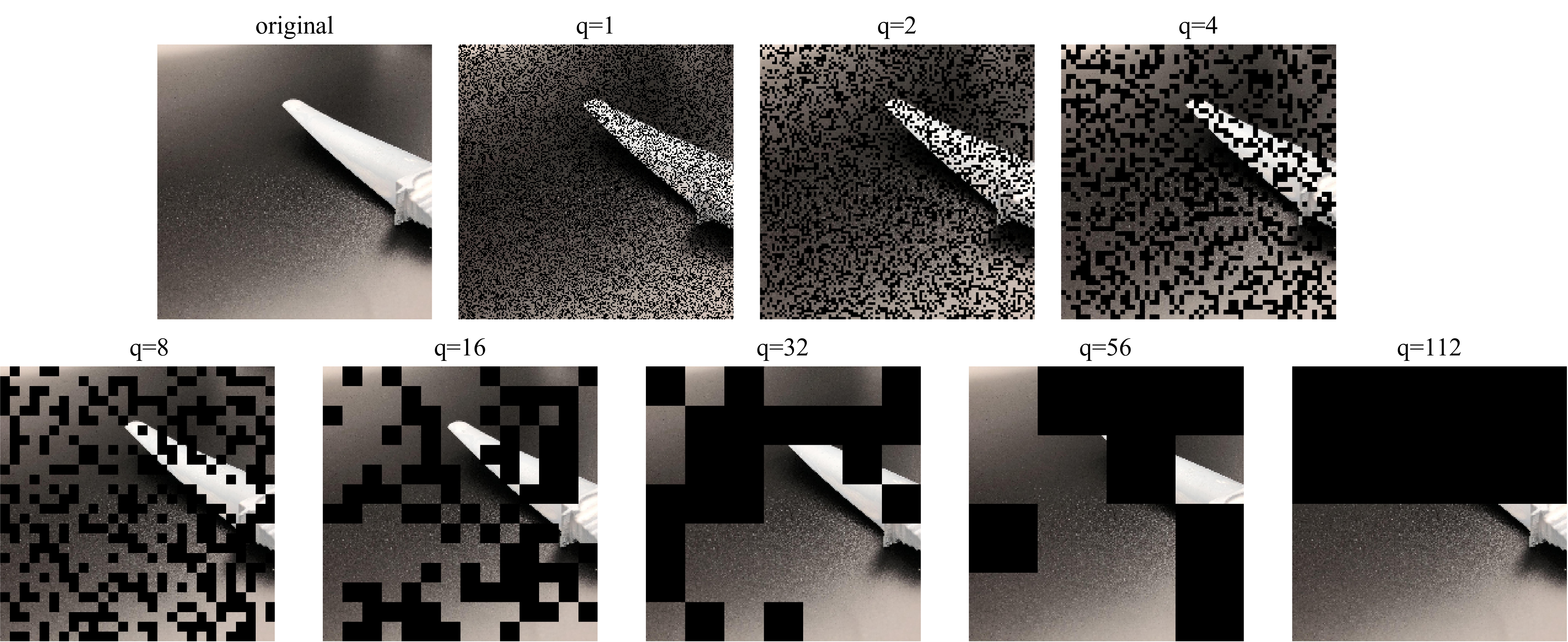}
\caption{Different forms of input with the same masking ratio. $q$ refers to the smallest unit ($q \times q$ pixels) of masking.}
\label{fig_abl_1_1}
\end{figure}

\begin{figure}[!t]
\centering
\includegraphics[width=3in]{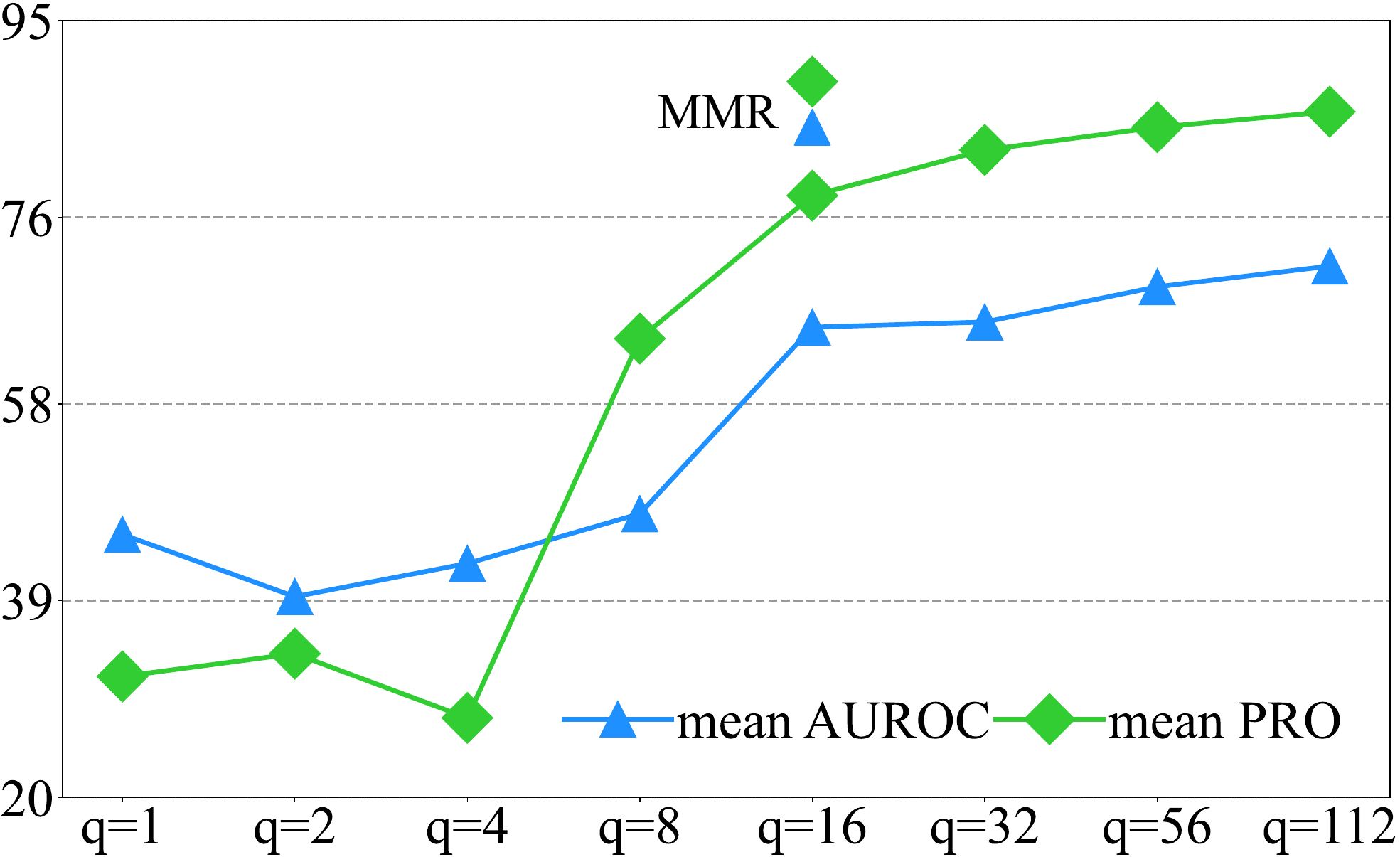}
\caption{Performance (mean AUROC (\%) and mean PRO (\%)) of different smallest units on AeBAD-S dataset.}
\label{fig_abl_1_2}
\end{figure}

\section{Ablation Study}
\label{6_abl}

In this part, we investigate some factors that have greater impacts on MMR. We choose AeBAD-S dataset for ablation study. If there is no additional statement, the setups and hyper-parameters of the experiments remain the same as in Section \ref{sec_id}.

\subsection{Form of Input}

\begin{figure}[!t]
\centering
\includegraphics[width=3in]{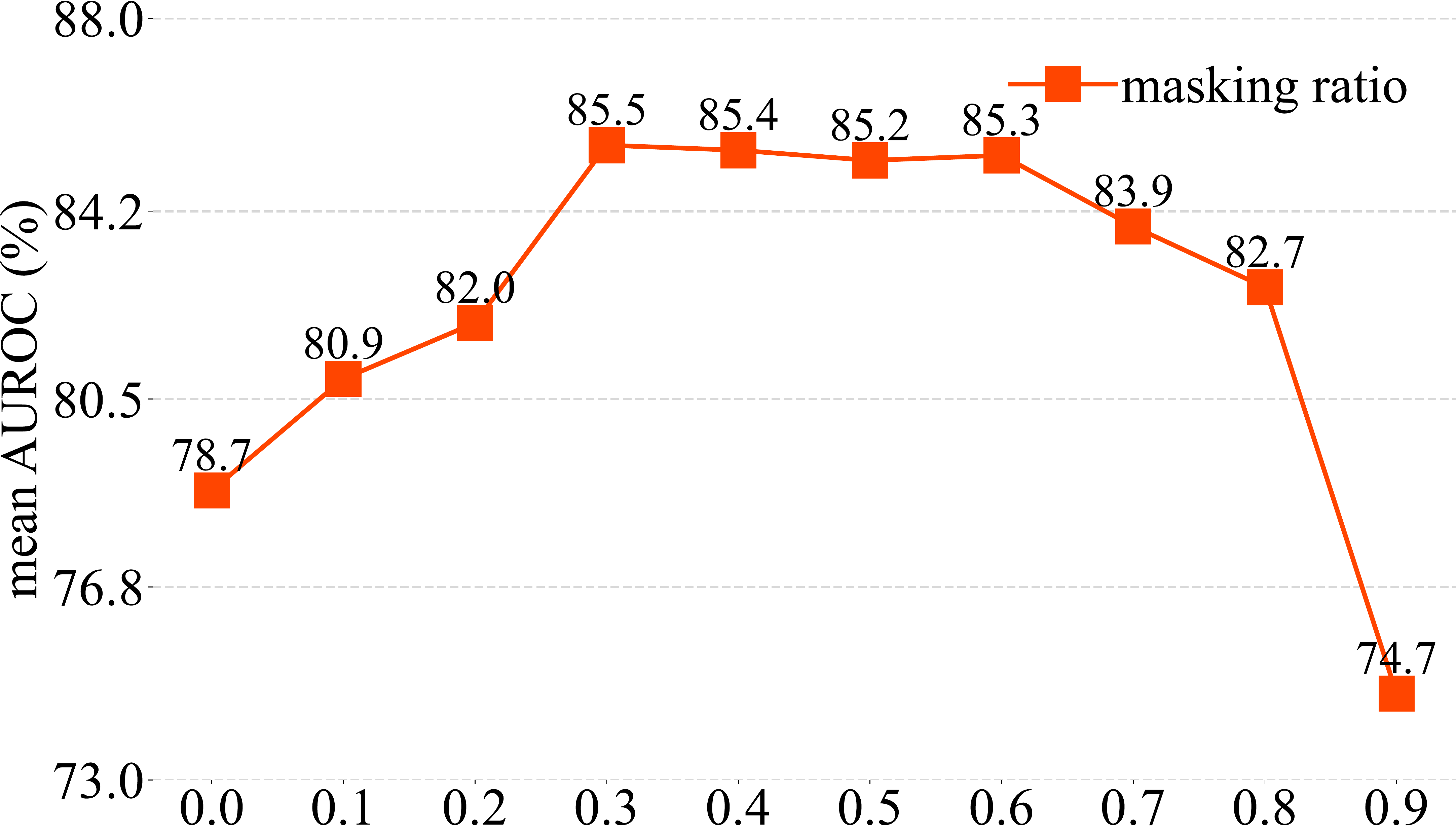}
\caption{Mean AUROC (\%) of different masking ratios on AeBAD-S dataset.}
\label{fig_abl_2}
\end{figure}

In MMR, we take non-masked patches as input. In this part, we select the entire image (all patches) as input while maintaining a constant masking ratio. We regard $q \times q$ pixels as a unit and then randomly mask $\eta=0.4$ times the area of the entire image based on this unit. We choose $q=[1, 2, 4, 8, 16, 32, 56, 112]$. An example of masked images of different $q$ is shown in Figure \ref{fig_abl_1_1}. Note that when $q=16$, the information included in the non-masked input is the same as the original input of MMR. The only difference is that the information in the non-masked part is shared, which is shown in Figure \ref{fig_mmr} (c). The results of different inputs are shown in Figure \ref{fig_abl_1_2}. We can observe that when the masking ratio is constant, the input with different masking forms can have a significant impact on performance. When $q\leqslant8$, the performance is poor. The reason is that although the information of the masked patches is unknown, the adjacent region will leak information due to the small unit. This makes the model lose the ability to infer the information in the masked patches from the global context. When $q=16$, the performance improves due to the larger unit. However, since this still cannot prevent the leakage of information, the performance at $q=16$ is lower than the proposed MMR at $q=16$. In addition, we find an interesting phenomenon: as the smallest unit continues to increase, performance will continue to improve. We believe that the main reason is that when the smallest unit becomes larger, the association between image blocks will become weaker and weaker, even if the information leaks, it will not affect the overall reconstruction. However, the details of some reconstructions will become worse, which leads to a large increase in its mean PRO and basically no growth in mean AUROC. This is similar to 0.9 mask ratio in Figure \ref{fig_abl_2}.

\subsection{Masking Ratio}

We conduct experiments with different masking ratios in order to explore how they affect the model's ability to perceive causal relationships. Figure \ref{fig_abl_2} displays the results of this analysis for masking ratios ranging from 0 to 0.9. Our findings suggest that the model's performance exhibits a pattern of low scores at both ends of the masking ratio, with the highest scores achieved in the middle. When the masking ratio is small, the model is able to make accurate inferences about the masked patches based on the surrounding regions that contain a lot of relevant information. However, as the masking ratio is high, the model becomes less able to infer the whole image based on just a few visible patches, resulting in lower performance scores. Note that different from the recent masked autoencoder \cite{he2022masked}, where the optimal masking ratio is high and the target is to perceive the semantic, the model for IAD needs to discriminate some tiny defects, and thus requires a smaller optimal masking ratio to reconstruct the details accurately.

\begin{table}
\scriptsize
    \centering
    \caption{Performance of pre-training of different types. PIMAE denotes that ViT backbone is pre-trained on ImageNet by a masked autoencoder.}
    \begin{tabular}{c c c}
    \hline
    & mean AUROC (\%) & mean PRO (\%) \\ \hline
    Train from scratch & 75.1 & 88.5 \\
    PIMAE & 84.7 & 89.1 \\ \hline
    \end{tabular}
    \label{ta_abl_1}
\end{table}

\subsection{Tpye of Pre-training}

In our implementation, the ViT backbone is pre-trained on ImageNet by a masked autoencoder \cite{he2022masked}. In this part, we show the result that ViT is trained from scratch in Table \ref{ta_abl_1}. Intuitively, training ViT from scratch using normal samples should result in a model that is better able to identify unseen anomalies. However, we find that this is not the case. The main reason for this is that training ViT on a small dataset from scratch can lead to overfitting, which in turn reduces the generalization ability of the model. In addition, another advantage of choosing ViT pre-trained by a masked autoencoder is that the pre-trained upstream tasks are roughly the same as our proposed downstream task, both of which aim to reconstruct the original input. This similarity allows the model to transfer its learned representations more effectively, resulting in better performance on our task.

\begin{table}
\scriptsize
    \centering
    \caption{Performance of different pre-trained hierarchical encoders. The pre-trained encoder is pre-training on ImageNet 1K.}
    \begin{tabular}{c c c}
    \hline
    & mean AUROC (\%) & mean PRO (\%) \\ \hline
    Random Initialized ResNet 18 & 51.8 & 52.8 \\
    Pre-trained ResNet 18 & 82.4 & 90.7 \\ \hline
    WideResNet 50 & 84.7 & 89.1 \\ \hline
    \end{tabular}
    \label{ta_abl_2}
\end{table}


\subsection{Frozen Pre-trained Hierarchical Encoder}

In this part, we choose different frozen pre-trained hierarchical encoders to train MMR. The results are shown in Table \ref{ta_abl_2}. We can observe that MMR can also generalize well to ResNet 18.

\begin{table}
\scriptsize
    \centering
    \caption{Effect of features at different scales on performance (Mean AUROC (\%) and Mean PRO (\%)).}
    \begin{tabular}{c c c c c}
    \hline
     & layer1 & layer2 & layer3 & Mean AUROC (\%) / Mean PRO (\%) \\ \hline
   1 & $\surd$ & $\ $ & $\ $ & 66.5 / 85.6 \\
   2  & $\ $ & $\surd $ & $\ $ & 76.9 / 86.6 \\ 
   3  & $\ $ & $\ $ & $\surd $ & 76.5 / 82.4 \\ 
   4 & $\surd$ & $\surd $ & $\ $ & 74.7 / 86.9 \\
   5  & $\ $ & $\surd $ & $\surd $ & \textbf{85.0} / 88.1 \\ 
   6  & $\surd $ & $\surd $ & $\surd $ & 84.7 / \textbf{89.1}\\ \hline
    \end{tabular}
    \label{Ta_8}
\end{table}

\subsection{Multi-scale Features}

In this part, we report the effect of features at different scales on performance. The results are shown in Table \ref{Ta_8}. We can observe that the features at the intermediate scale are more conducive to detecting defects, while the features at the other scales can be a complement to improve performance.

\begin{table}
\scriptsize
    \centering
    \caption{Throughput (image/s) of current SOTA methods and MMR.}
    \begin{tabular}{c c c c}
    \hline
    & PatchCore \cite{roth2022towards} & RD \cite{deng2022anomaly} & MMR (Ours) \\ \hline
    Throughput (image/s) & 76.0 & \textbf{190.4} & 118.1 \\ \hline
    \end{tabular}
    \label{ta_throughput}
\end{table}

\section{Limitations and Future Works}
\label{7_lim}

The main limitation of MMR is that it needs two base models (a frozen pre-trained encoder and a specific trained model) to infer the test image, which consumes a lot of memory. This limits its usage in practical application. In addition, the proposed causal transformer-based module is memory-intensive, which also hinders practical applications. We calculate the throughput of some SOTA models and MMR, where the experimental condition is described in Section \ref{sec_id} and the input image is $224 \times 224$. The results are listed in Table \ref{ta_throughput}. We can observe that since there are two base models in MMR, its throughput is inferior to ReverseDistillation. In future work, we will explore the causal module based on the convolution module and integrate it with the pre-trained model in the single model.

Furthermore, the proposed dataset only focuses on a single object (blade), which cannot fully show the capability of anomaly detection of the model. We will explore more practical scenarios of anomaly detection, which will include more defective types.

\section{Conclusion}
\label{7_con}

The aim of this paper is to introduce AeBAD, a new dataset that addresses the issue of domain shift, which has not been fully considered in existing datasets. Through our analysis on this dataset, we have identified deficiencies in current SOTA methods when the domain of normal samples in the test set shifts. We have also investigated the underlying causes of these deficiencies and proposed a new method called MMR, which enhances the model's ability to infer causality among patches in normal samples and improves performance on different domain shifts. Moreover, MMR is competitive in detecting various types of anomalies. Our work presents a new direction for the anomaly detection community, one that is more closely aligned with real-world industrial scenarios.

\section*{Acknowledgments}
This work is supported by National Natural Science Foundation of China under Grant 52175114\&92060302, Science Center for Gas Turbine Project (P2022-DC-I-003-001) and Special support plan for high level talents in Shaanxi Province..

{\small
\bibliographystyle{ieee_fullname}
\bibliography{ref}

\begin{thebibliography}{10}\itemsep=-1pt

\bibitem{adelson1984pyramid}
Edward~H Adelson, Charles~H Anderson, James~R Bergen, Peter~J Burt, and Joan~M
  Ogden.
\newblock Pyramid methods in image processing.
\newblock {\em RCA engineer}, 29(6):33--41, 1984.

\bibitem{ba2016layer}
Jimmy~Lei Ba, Jamie~Ryan Kiros, and Geoffrey~E Hinton.
\newblock Layer normalization.
\newblock {\em arXiv preprint arXiv:1607.06450}, 2016.

\bibitem{bergmann2022beyond}
Paul Bergmann, Kilian Batzner, Michael Fauser, David Sattlegger, and Carsten
  Steger.
\newblock Beyond dents and scratches: Logical constraints in unsupervised
  anomaly detection and localization.
\newblock {\em International Journal of Computer Vision}, 130(4):947--969,
  2022.

\bibitem{bergmann2019mvtec}
Paul Bergmann, Michael Fauser, David Sattlegger, and Carsten Steger.
\newblock Mvtec ad--a comprehensive real-world dataset for unsupervised anomaly
  detection.
\newblock In {\em Proceedings of the IEEE/CVF Conference on Computer Vision and
  Pattern Recognition}, pages 9592--9600, 2019.

\bibitem{bergmann2020uninformed}
Paul Bergmann, Michael Fauser, David Sattlegger, and Carsten Steger.
\newblock Uninformed students: Student-teacher anomaly detection with
  discriminative latent embeddings.
\newblock In {\em Proceedings of the IEEE/CVF Conference on Computer Vision and
  Pattern Recognition}, pages 4183--4192, 2020.

\bibitem{bergmann2021mvtec}
Paul Bergmann, Xin Jin, David Sattlegger, and Carsten Steger.
\newblock The mvtec 3d-ad dataset for unsupervised 3d anomaly detection and
  localization.
\newblock {\em arXiv preprint arXiv:2112.09045}, 2021.

\bibitem{bergmann2018improving}
Paul Bergmann, Sindy L{\"o}we, Michael Fauser, David Sattlegger, and Carsten
  Steger.
\newblock Improving unsupervised defect segmentation by applying structural
  similarity to autoencoders.
\newblock {\em arXiv preprint arXiv:1807.02011}, 2018.

\bibitem{bhardwaj2023steerable}
Sangnie Bhardwaj, Willie McClinton, Tongzhou Wang, Guillaume Lajoie, Chen Sun,
  Phillip Isola, and Dilip Krishnan.
\newblock Steerable equivariant representation learning.
\newblock {\em arXiv preprint arXiv:2302.11349}, 2023.

\bibitem{cui2022survey}
Yajie Cui, Zhaoxiang Liu, and Shiguo Lian.
\newblock A survey on unsupervised industrial anomaly detection algorithms.
\newblock {\em arXiv preprint arXiv:2204.11161}, 2022.

\bibitem{defard2020padim}
Thomas Defard, Aleksandr Setkov, Angelique Loesch, and Romaric Audigier.
\newblock Padim: a patch distribution modeling framework for anomaly detection
  and localization.
\newblock {\em arXiv preprint arXiv:2011.08785}, 2020.

\bibitem{dehaene2020iterative}
David Dehaene, Oriel Frigo, S{\'e}bastien Combrexelle, and Pierre Eline.
\newblock Iterative energy-based projection on a normal data manifold for
  anomaly localization.
\newblock {\em arXiv preprint arXiv:2002.03734}, 2020.

\bibitem{deng2022anomaly}
Hanqiu Deng and Xingyu Li.
\newblock Anomaly detection via reverse distillation from one-class embedding.
\newblock In {\em Proceedings of the IEEE/CVF Conference on Computer Vision and
  Pattern Recognition}, pages 9737--9746, 2022.

\bibitem{dosovitskiy2020image}
Alexey Dosovitskiy, Lucas Beyer, Alexander Kolesnikov, Dirk Weissenborn,
  Xiaohua Zhai, Thomas Unterthiner, Mostafa Dehghani, Matthias Minderer, Georg
  Heigold, Sylvain Gelly, et~al.
\newblock An image is worth 16x16 words: Transformers for image recognition at
  scale.
\newblock In {\em International Conference on Learning Representations}, 2020.

\bibitem{gudovskiy2022cflow}
Denis Gudovskiy, Shun Ishizaka, and Kazuki Kozuka.
\newblock Cflow-ad: Real-time unsupervised anomaly detection with localization
  via conditional normalizing flows.
\newblock In {\em Proceedings of the IEEE/CVF Winter Conference on Applications
  of Computer Vision}, pages 98--107, 2022.

\bibitem{he2022masked}
Kaiming He, Xinlei Chen, Saining Xie, Yanghao Li, Piotr Doll{\'a}r, and Ross
  Girshick.
\newblock Masked autoencoders are scalable vision learners.
\newblock In {\em Proceedings of the IEEE/CVF Conference on Computer Vision and
  Pattern Recognition}, pages 16000--16009, 2022.

\bibitem{he2016deep}
Kaiming He, Xiangyu Zhang, Shaoqing Ren, and Jian Sun.
\newblock Deep residual learning for image recognition.
\newblock In {\em Proceedings of the IEEE conference on computer vision and
  pattern recognition}, pages 770--778, 2016.

\bibitem{hendrycks2016gaussian}
Dan Hendrycks and Kevin Gimpel.
\newblock Gaussian error linear units (gelus).
\newblock {\em arXiv preprint arXiv:1606.08415}, 2016.

\bibitem{hinton2015distilling}
Geoffrey Hinton, Oriol Vinyals, and Jeff Dean.
\newblock Distilling the knowledge in a neural network.
\newblock {\em arXiv preprint arXiv:1503.02531}, 2015.

\bibitem{8963630}
Jian Huang, Junzhe Wang, Yihua Tan, Dongrui Wu, and Yu Cao.
\newblock An automatic analog instrument reading system using computer vision
  and inspection robot.
\newblock {\em IEEE Transactions on Instrumentation and Measurement},
  69(9):6322--6335, 2020.

\bibitem{jiang2022survey}
Xi Jiang, Guoyang Xie, Jinbao Wang, Yong Liu, Chengjie Wang, Feng Zheng, and
  Yaochu Jin.
\newblock A survey of visual sensory anomaly detection.
\newblock {\em arXiv preprint arXiv:2202.07006}, 2022.

\bibitem{li2021cutpaste}
Chun-Liang Li, Kihyuk Sohn, Jinsung Yoon, and Tomas Pfister.
\newblock Cutpaste: Self-supervised learning for anomaly detection and
  localization.
\newblock In {\em Proceedings of the IEEE/CVF Conference on Computer Vision and
  Pattern Recognition}, pages 9664--9674, 2021.

\bibitem{9363206}
Dawei Li, Yida Li, Qian Xie, Yuxiang Wu, Zhenghao Yu, and Jun Wang.
\newblock Tiny defect detection in high-resolution aero-engine blade images via
  a coarse-to-fine framework.
\newblock {\em IEEE Transactions on Instrumentation and Measurement}, 70:1--12,
  2021.

\bibitem{li2022exploring}
Yanghao Li, Hanzi Mao, Ross Girshick, and Kaiming He.
\newblock Exploring plain vision transformer backbones for object detection.
\newblock In {\em Computer Vision--ECCV 2022: 17th European Conference, Tel
  Aviv, Israel, October 23--27, 2022, Proceedings, Part IX}, pages 280--296.
  Springer, 2022.

\bibitem{lin2017feature}
Tsung-Yi Lin, Piotr Doll{\'a}r, Ross Girshick, Kaiming He, Bharath Hariharan,
  and Serge Belongie.
\newblock Feature pyramid networks for object detection.
\newblock In {\em Proceedings of the IEEE conference on computer vision and
  pattern recognition}, pages 2117--2125, 2017.

\bibitem{liu2023learning}
Ye Liu, Jun Chen, and Jia-ao Hou.
\newblock Learning position information from attention: End-to-end weakly
  supervised crack segmentation with gans.
\newblock {\em Computers in Industry}, 149:103921, 2023.

\bibitem{loshchilov2017decoupled}
Ilya Loshchilov and Frank Hutter.
\newblock Decoupled weight decay regularization.
\newblock {\em arXiv preprint arXiv:1711.05101}, 2017.

\bibitem{matsubara2020deep}
Takashi Matsubara, Kazuki Sato, Kenta Hama, Ryosuke Tachibana, and Kuniaki
  Uehara.
\newblock Deep generative model using unregularized score for anomaly detection
  with heterogeneous complexity.
\newblock {\em IEEE Transactions on Cybernetics}, 52(6):5161--5173, 2020.

\bibitem{nag2022wafersegclassnet}
Subhrajit Nag, Dhruv Makwana, Sparsh Mittal, C~Krishna Mohan, et~al.
\newblock Wafersegclassnet-a light-weight network for classification and
  segmentation of semiconductor wafer defects.
\newblock {\em Computers in Industry}, 142:103720, 2022.

\bibitem{9927464}
Christopher Naverschnigg, Ernst Csencsics, and Georg Schitter.
\newblock Flexible robot-based in-line measurement system for high-precision
  optical surface inspection.
\newblock {\em IEEE Transactions on Instrumentation and Measurement}, 71:1--9,
  2022.

\bibitem{pirnay2022inpainting}
Jonathan Pirnay and Keng Chai.
\newblock Inpainting transformer for anomaly detection.
\newblock In {\em Image Analysis and Processing--ICIAP 2022: 21st International
  Conference, Lecce, Italy, May 23--27, 2022, Proceedings, Part II}, pages
  394--406. Springer, 2022.

\bibitem{ravcki2022detection}
Domen Ra{\v{c}}ki, Dejan Toma{\v{z}}evi{\v{c}}, and Danijel Sko{\v{c}}aj.
\newblock Detection of surface defects on pharmaceutical solid oral dosage
  forms with convolutional neural networks.
\newblock {\em Neural Computing and Applications}, 34(1):631--650, 2022.

\bibitem{rahman2023railway}
Miftahur Rahman, Haochen Liu, Mohammed Masri, Isidro Durazo-Cardenas, and
  Andrew Starr.
\newblock A railway track reconstruction method using robotic vision on a
  mobile manipulator: A proposed strategy.
\newblock {\em Computers in Industry}, 148:103900, 2023.

\bibitem{roth2022towards}
Karsten Roth, Latha Pemula, Joaquin Zepeda, Bernhard Sch{\"o}lkopf, Thomas
  Brox, and Peter Gehler.
\newblock Towards total recall in industrial anomaly detection.
\newblock In {\em Proceedings of the IEEE/CVF Conference on Computer Vision and
  Pattern Recognition}, pages 14318--14328, 2022.

\bibitem{russakovsky2015imagenet}
Olga Russakovsky, Jia Deng, Hao Su, Jonathan Krause, Sanjeev Satheesh, Sean Ma,
  Zhiheng Huang, Andrej Karpathy, Aditya Khosla, Michael Bernstein, et~al.
\newblock Imagenet large scale visual recognition challenge.
\newblock {\em International journal of computer vision}, 115(3):211--252,
  2015.

\bibitem{salehi2021multiresolution}
Mohammadreza Salehi, Niousha Sadjadi, Soroosh Baselizadeh, Mohammad~H Rohban,
  and Hamid~R Rabiee.
\newblock Multiresolution knowledge distillation for anomaly detection.
\newblock In {\em Proceedings of the IEEE/CVF conference on computer vision and
  pattern recognition}, pages 14902--14912, 2021.

\bibitem{schluter2022natural}
Hannah~M Schl{\"u}ter, Jeremy Tan, Benjamin Hou, and Bernhard Kainz.
\newblock Natural synthetic anomalies for self-supervised anomaly detection and
  localization.
\newblock In {\em Computer Vision--ECCV 2022: 17th European Conference, Tel
  Aviv, Israel, October 23--27, 2022, Proceedings, Part XXXI}, pages 474--489.
  Springer, 2022.

\bibitem{shi2023few}
Xiangwen Shi, Shaobing Zhang, Miao Cheng, Lian He, Xianghong Tang, and Zhe Cui.
\newblock Few-shot semantic segmentation for industrial defect recognition.
\newblock {\em Computers in Industry}, 148:103901, 2023.

\bibitem{tabernik2020segmentation}
Domen Tabernik, Samo {\v{S}}ela, Jure Skvar{\v{c}}, and Danijel Sko{\v{c}}aj.
\newblock Segmentation-based deep-learning approach for surface-defect
  detection.
\newblock {\em Journal of Intelligent Manufacturing}, 31(3):759--776, 2020.

\bibitem{tao2022deep}
Xian Tao, Xinyi Gong, Xin Zhang, Shaohua Yan, and Chandranath Adak.
\newblock Deep learning for unsupervised anomaly localization in industrial
  images: A survey.
\newblock {\em IEEE Transactions on Instrumentation and Measurement}, 2022.

\bibitem{wieler2007weakly}
Matthias Wieler and Tobias Hahn.
\newblock Weakly supervised learning for industrial optical inspection.
\newblock In {\em DAGM symposium in}, 2007.

\bibitem{xisoftpatch}
Jiang Xi, Jianlin Liu, Jinbao Wang, Qiang Nie, WU Kai, Yong Liu, Chengjie Wang,
  and Feng Zheng.
\newblock Softpatch: Unsupervised anomaly detection with noisy data.
\newblock In {\em Advances in Neural Information Processing Systems}.

\bibitem{yang2020dfr}
Jie Yang, Yong Shi, and Zhiquan Qi.
\newblock Dfr: Deep feature reconstruction for unsupervised anomaly
  segmentation.
\newblock {\em arXiv preprint arXiv:2012.07122}, 2020.

\bibitem{yang2022review}
Pingping Yang, Wenhui Yue, Jian Li, Guangfu Bin, and Chao Li.
\newblock Review of damage mechanism and protection of aero-engine blades based
  on impact properties.
\newblock {\em Engineering Failure Analysis}, page 106570, 2022.

\bibitem{you2022unified}
Zhiyuan You, Lei Cui, Yujun Shen, Kai Yang, Xin Lu, Yu Zheng, and Xinyi Le.
\newblock A unified model for multi-class anomaly detection.
\newblock {\em arXiv preprint arXiv:2206.03687}, 2022.

\bibitem{yu2021fastflow}
Jiawei Yu, Ye Zheng, Xiang Wang, Wei Li, Yushuang Wu, Rui Zhao, and Liwei Wu.
\newblock Fastflow: Unsupervised anomaly detection and localization via 2d
  normalizing flows.
\newblock {\em arXiv preprint arXiv:2111.07677}, 2021.

\bibitem{zagoruyko2016wide}
Sergey Zagoruyko and Nikos Komodakis.
\newblock Wide residual networks.
\newblock {\em arXiv preprint arXiv:1605.07146}, 2016.

\bibitem{zavrtanik2021draem}
Vitjan Zavrtanik, Matej Kristan, and Danijel Sko{\v{c}}aj.
\newblock Draem-a discriminatively trained reconstruction embedding for surface
  anomaly detection.
\newblock In {\em Proceedings of the IEEE/CVF International Conference on
  Computer Vision}, pages 8330--8339, 2021.

\bibitem{zavrtanik2021reconstruction}
Vitjan Zavrtanik, Matej Kristan, and Danijel Sko{\v{c}}aj.
\newblock Reconstruction by inpainting for visual anomaly detection.
\newblock {\em Pattern Recognition}, 112:107706, 2021.

\bibitem{zeiser2023evaluation}
Alexander Zeiser, Bekir {\"O}zcan, Bas van Stein, and Thomas B{\"a}ck.
\newblock Evaluation of deep unsupervised anomaly detection methods with a
  data-centric approach for on-line inspection.
\newblock {\em Computers in Industry}, 146:103852, 2023.

\bibitem{zhang2022nico++}
Xingxuan Zhang, Linjun Zhou, Renzhe Xu, Peng Cui, Zheyan Shen, and Haoxin Liu.
\newblock Nico++: Towards better benchmarking for domain generalization.
\newblock {\em arXiv preprint arXiv:2204.08040}, 2022.

\bibitem{zhao2022ood}
Bingchen Zhao, Shaozuo Yu, Wufei Ma, Mingxin Yu, Shenxiao Mei, Angtian Wang, Ju
  He, Alan Yuille, and Adam Kortylewski.
\newblock Ood-cv: A benchmark for robustness to out-of-distribution shifts of
  individual nuisances in natural images.
\newblock In {\em Computer Vision--ECCV 2022: 17th European Conference, Tel
  Aviv, Israel, October 23--27, 2022, Proceedings, Part VIII}, pages 163--180.
  Springer, 2022.

\bibitem{zou2022spot}
Yang Zou, Jongheon Jeong, Latha Pemula, Dongqing Zhang, and Onkar Dabeer.
\newblock Spot-the-difference self-supervised pre-training for anomaly
  detection and segmentation.
\newblock In {\em Computer Vision--ECCV 2022: 17th European Conference, Tel
  Aviv, Israel, October 23--27, 2022, Proceedings, Part XXX}, pages 392--408.
  Springer, 2022.

\end{thebibliography}
}

\end{document}